\documentclass[10pt,twocolumn,letterpaper]{article}

\usepackage{cvpr}      
\usepackage{xcolor}

\usepackage{subcaption}
\usepackage{tcolorbox}
\usepackage{scalerel}
\usepackage{makecell}
\usepackage{amsmath}
\usepackage{multirow}
\usepackage{colortbl, xcolor}
\newtcbox{\roundhighlightyellow}{on line,
  arc=3pt, colframe=yellow, colback=yellow!30, boxrule=0pt, boxsep=0pt, left=3pt, right=3pt, top=2pt, bottom=2pt}

\definecolor{softgreen}{RGB}{208, 236, 227}
\newtcbox{\roundhighlightgreen}{on line,
  arc=3pt, colframe=green, colback=softgreen, boxrule=0pt, boxsep=0pt, left=3pt, right=3pt, top=2pt, bottom=2pt}

\definecolor{teal}{RGB}{178, 228, 211}
\newtcbox{\roundhighlightteal}{on line,
  arc=3pt, colframe=teal, colback=teal, boxrule=0pt, boxsep=0pt, left=3pt, right=3pt, top=2pt, bottom=2pt}

  \definecolor{orange}{RGB}{252, 204, 184}
\newtcbox{\roundhighlightorange}{on line,
  arc=3pt, colframe=orange, colback=orange, boxrule=0pt, boxsep=0pt, left=3pt, right=3pt, top=2pt, bottom=2pt}

\newcommand{\graytext}[1]{\textcolor{gray}{#1}}

\definecolor{cvprblue}{rgb}{0.21,0.49,0.74}
\usepackage[pagebackref,breaklinks,colorlinks,allcolors=cvprblue]{hyperref}
\usepackage{bm}

\def\paperID{5861} 
\def\confName{ICCV}
\def\confYear{2025}

\title{Recovering Parametric Scenes from Very Few Time-of-Flight Pixels}

\author{Carter Sifferman\footnotemark[2], Yiquan Li\footnotemark[2], Yiming Li, Fangzhou Mu, Michael Gleicher, Mohit Gupta, Yin Li\\ 
University of Wisconsin-Madison
}

\newsavebox{\myglyph}
\sbox{\myglyph}{\includegraphics{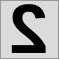}}

\begin{document}
\twocolumn[{
    \renewcommand\twocolumn[1][]{#1}
    \vspace{-1em}
    \maketitle
    \vspace{-3em}
    \begin{center}
        \centering
        \includegraphics[width=0.95\linewidth]{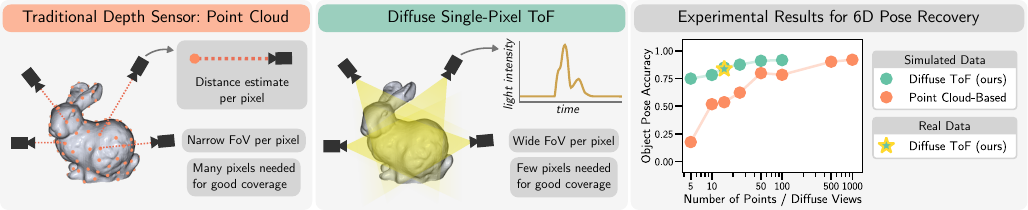}
        \vspace{-0.8em}
        \captionof{figure}{We introduce a method for recovering the geometry of parametric $3$D scenes, such as the $6$D pose of a known object, from a distributed set of \textbf{very few} (\eg, $15$), \roundhighlightteal{diffuse} (\ie, wide field-of-view) single-pixel ToF Sensors. Methods based on \roundhighlightorange{traditional} depth sensors suffer poor performance under a low-pixel-count regime due to their sparse coverage. Our method outperforms a point cloud-based baseline by utilizing the entirety of data recovered by a diffuse ToF sensor. See \cref{subsec:main_results} for details of the experiment.
        }\vspace{-0.5em}
        \label{fig:teaser}
    \end{center}
}]

{
  \renewcommand{\thefootnote}
    {\fnsymbol{footnote}}
  \footnotetext[2]{Co-first author}
}

\begin{abstract}

We aim to recover the geometry of 3D parametric scenes using \textbf{very few} depth measurements from low-cost, commercially available time-of-flight sensors. These sensors offer very low spatial resolution (i.e., a single pixel), but image a wide field-of-view per pixel and capture detailed time-of-flight data in the form of time-resolved photon counts. This time-of-flight data encodes rich scene information and thus enables recovery of simple scenes from sparse measurements. We investigate the feasibility of using a distributed set of few measurements (e.g., as few as 15 pixels) to recover the geometry of simple parametric scenes with a strong prior, such as estimating the 6D pose of a known object. To achieve this, we design a method that utilizes both feed-forward prediction to infer scene parameters, and differentiable rendering within an analysis-by-synthesis framework to refine the scene parameter estimate. We develop hardware prototypes and demonstrate that our method effectively recovers object pose given an untextured 3D model in both simulations and controlled real-world captures, and show promising initial results for other parametric scenes. We additionally conduct experiments to explore the limits and capabilities of our imaging solution. Our project webpage is available at  \href{https://cpsiff.github.io/recovering_parametric_scenes}{\mbox{cpsiff.github.io/recovering\_parametric\_scenes}}

\end{abstract}    
\section{Introduction}
\label{sec:intro}

Time-of-flight (ToF) cameras such as LiDARs are a key technology for modern 3D vision, enabling tasks like pose estimation, shape reconstruction, and object recognition, with applications spanning fields such as robotics, augmented reality, and autonomous driving. Most current methods for inference from ToF imagery depend on dense 3D data, usually represented as a point cloud captured by high-resolution camera(s). It is generally accepted that a dense collection of depth measurements (\eg, ToF pixels as points) is vital for precise 3D vision. While certain applications \emph{do require} high-resolution geometry, can some vision tasks be accomplished with only sparse 3D measurements? 


This question is particularly relevant given the recent emergence of low-cost ($<\$3$ USD per unit), miniature ($<5$ mm across) ToF sensors~\cite{VL6180X_, TMF8820_}. These sensors, already deployed in mobile~\cite{techinsights2023iphone} and wearable applications~\cite{fierce2025meta}, are often implemented with a single photon avalanche diode (SPAD) array~\cite{niclass_design_2005, morimoto_megapixel_2020}, featuring very limited pixel counts (even a single pixel) yet a wide field-of-view (FoV) per pixel (\eg, $30$\textdegree). They capture raw ToF data with a \textit{transient histogram}---a $1$D waveform that encodes the intensity of light returning from the scene at pico-to-nanosecond timescales, integrated over a pixel's entire FoV. Traditionally, these histograms are processed into a point cloud by detecting and converting their peaks into depth estimates. However, this processing reduces the high-dimensional histogram to a single number, eliminating potentially useful information. Our key hypothesis, inspired by recent studies~\cite{mu_towards_2024,behari_blurred_2024,avidan_deltar_2022,avidan_3d_2022}, is that while these sensors cannot recover high-resolution point clouds, even a few transient histograms encode rich scene information sufficient for various downstream 3D vision tasks. An example is recovering scenes with low geometric complexity or scenarios where a strong geometric prior (\eg, a low-dimensional parametric shape model) is available. Our question is, under these conditions, \textit{what is the minimal number of depth measurements required to recover 3D scenes?} 



As a first step towards addressing this question, we investigate the recovery of simple parametric 3D scenes using \textit{very few} ToF measurements, each with a wide FoV (see~\cref{fig:teaser}), \eg, as few as $15$ pixels captured by spatially distributed, low-cost single-pixel SPAD sensors. We assume a 3D Lambertian scene defined by a parametric shape model and aim to recover the parameters of that model using a limited number of transient histograms captured from known fixed poses. We place special emphasis on the task of 6D pose estimation, which is a specific case of parametric scene recovery. In this case, the parameters that we aim to recover are the position and orientation of a known object mesh. We focus on 6D pose estimation because it is a well-defined problem with practical applications in robotics and augmented reality. This makes it a good testbed to tackle the fundamental challenge: utilizing very low resolution sensor data. With very few pixels, each of which integrates over a wide FoV, recovering 6D pose is challenging.

To solve this problem, we present an \textit{analysis-by-synthesis} based approach. Our method integrates (1) a learning-based feedforward model which predicts an initial estimate of scene parameters; (2) a differentiable renderer that synthesizes sensor measurements given scene parameters in the parametric model; and (3) an optimization-based refiner that iteratively renders sensor measurements to optimize scene parameters using the differentiable renderer. To address the scarcity of real-world imaging data, we re-use our renderer to generate a large-scale synthetic dataset for training our feedforward model, and explore its ability to transfer to real-world captures.

%
We develop hardware prototypes for real-world capture and evaluate our approach using both simulated and real-world data. In real-world tests, our approach successfully estimates poses of even non-Lambertian objects using only $15$ ToF pixels and an untextured object mesh. Moreover, leveraging our approach and hardware, we also briefly investigate two other forms of parametric scene recovery: parametric shape recovery (\ie, the position and scale of a known spherical object), and human hand pose recovery (\ie, pose and articulation), for which we demonstrate encouraging preliminary results in real-world settings. 

\smallskip 

\noindent \textbf{Scope and Limitations}. While our problem formulation is general, we focus on 6D pose estimation, with a limited exploration of two other forms of parametric scene recovery. Our main objective is to establish feasibility rather than present an immediately deployable solution. To simplify real-world experiments, we make key assumptions, such as Lambertian surfaces, co-located sensor and light source, and known sensor poses. While robustness to varying scene reflectance and imperfect sensor poses are assessed in our experiments, our approach and prototype are not yet practical for widespread use. Moreover, we rely on currently available consumer hardware, which has a restricted sensing range, limiting our experiments to tabletop scenes.

\section{Related Work}
\label{sec:related_work}

\noindent \textbf{Time-of-flight (ToF) Imaging with SPADs}.
A ToF camera emits light pulses and measures the return time of incident photons to estimate distance. SPAD sensors have increasingly been adopted for ToF imaging, typically combined with a co-located light source (\eg, laser)~\cite{morimoto_megapixel_2020}. This setup has been successfully applied to fluorescence lifetime imaging~\cite{schwartz2008single}, novel view synthesis~\cite{malik2024flying}, and non-line-of-sight (NLOS) imaging~\cite{xin_theory_2019,heide2019non,klinghoffer2024platonerf,kirmani2009looking,faccio2020non}. Many of these systems rely on large, costly SPAD arrays ($>$\$10K USD) with high spatial and temporal resolution. Recent works have explored low-cost SPAD sensors ($<$\$3 USD) with limited pixel counts and lower temporal resolution for applications such as NLOS imaging~\cite{callenberg_low-cost_2021, young_enhancing_2024}, shape reconstruction~\cite{mu_towards_2024}, human pose estimation~\cite{ruget_pixels2pose_2022}, and SLAM~\cite{liu2023multi}. Our work also explores low-cost SPAD sensors for ToF imaging; however, our primary focus is on the feasibility of using a minimal number of SPAD sensors for parametric scene recovery.

A key component in SPAD imaging is modeling the image formation process. Sifferman \etal~\cite{sifferman_unlocking_2023} introduce a simple model for miniature ToF sensors. Recent works model the SPAD image formation process with differentiable functions for laboratory-grade~\cite{malik_transient_2023, luo_transientangelo_2024} or commodity~\cite{mu_towards_2024} sensors, enabling gradient-based optimization and facilitating 3D tasks such as pose estimation and shape reconstruction. Our work modifies the sensor model in~\cite{mu_towards_2024} to accommodate differentiable mesh rendering.
\smallskip

\begin{figure*}[th!]
    \centering
    \includegraphics[width=0.9\linewidth]{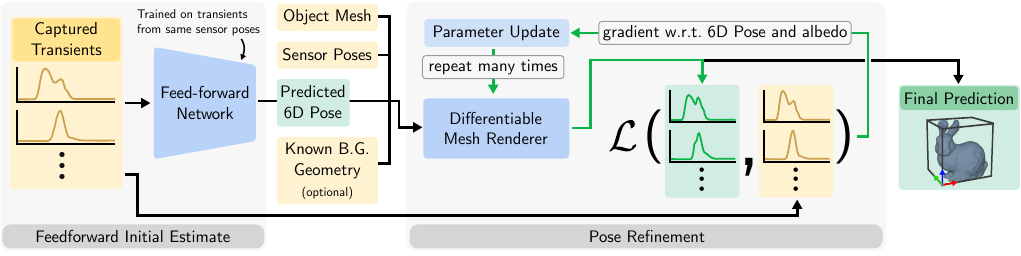}\vspace{-0.7em}
    \caption{\textbf{Overview of our method} as applied to 6D pose estimation. Our method consists of two components: 1) a pose prediction module, where a feedforward network estimates initial object pose from a sparse set of input transient histograms; and 2) a pose refiner, where a differentiable renderer is integrated into an analysis-by-synthesis framework to iteratively optimize the pose estimates. \roundhighlightyellow{Yellow~boxes} indicate inputs. \roundhighlightgreen{Green~boxes} indicate (intermediate) outputs. The optimization loop is illustrated with \textcolor{ForestGreen}{green arrows}.}\vspace{-1.5em}
    \label{fig:method_diagram}
\end{figure*}

\noindent \textbf{3D Vision with Low-Cost SPADs}. Prior works have explored feedforward neural networks for inference from transient histograms. Pixels2Pose~\cite{ruget_pixels2pose_2022} proposes a learning-based method to estimate whole-body human pose from a single 4$\times$4 pixel transient histogram. 
DELTAR~\cite{avidan_deltar_2022} and Jungerman \etal~\cite{avidan_3d_2022} both predict high resolution depth images from ToF transient(s) plus an RGB image. The neural networks in these prior works are generally trained on real-world data only, or on simulated data generated from a simple sensor model. In this work, we train on simulated data generated via a high-fidelity sensor model.

Recent works have considered an analysis-by-synthesis (AbS) paradigm, which uses differentiable rendering to align a set of underlying scene parameters with observed transient measurements. Sifferman \etal~\cite{sifferman_unlocking_2023} design an AbS pipeline to recover 3DoF plane pose and albedo from a set of 3$\times$3 transient measurements. Mu \etal~\cite{mu_towards_2024} present a method to recover arbitrary 3D scenes from a distributed set of $>$100 single-pixel transient measurements. Behari \etal~\cite{behari_blurred_2024} leverage AbS to reconstruct arbitrary 3D scenes from a miniature ToF sensor plus an RGB camera. Liu \etal~\cite{liu2023multi} build a neural radiance field for dense SLAM by integrating data from a miniature ToF sensor with an RGB camera. Luo \etal~\cite{luo_transientangelo_2024} reconstruct 3D scenes from transient measurements from very few viewpoints, however they use a high-resolution scanning LiDAR with higher fidelity and data rate than the miniature ToF sensors we consider. 

Our work shares a similar goal to prior studies that use multiple sensors for 3D scene recovery. However, our focus is on leveraging strong geometric priors to push the limits of scene recovery using only a few low-fidelity sensors.\smallskip


\noindent \textbf{6D Pose Estimation}.
6D pose estimation aims to determine the $6$ degree-of-freedom pose of a known rigid object relative to the camera, given its 3D mesh model.
Recent approaches use supervised learning to directly regress object pose from RGB and/or depth images~\cite{wen_foundationpose_2024, xiang_posecnn_2018, park_pix2pose_2019, labbe_megapose_2022, cai_ove6d_2022}. The predicted pose can be further refined via an AbS pipeline~\cite{wen_foundationpose_2024, tremblay_diff-dope_2023, labbe_megapose_2022, krull_learning_2015}. We take inspiration from the success of these works in designing our approach, integrating a feedforward network for initial pose prediction and an AbS pipeline for pose refinement.

\newcommand{\bx}{\bm{x}}
\newcommand{\bs}{\bm{s}}
\newcommand{\bl}{\bm{l}}
\newcommand{\ns}{\hat{\bm{s}}}
\newcommand{\bw}{\bm{\omega}}
\newcommand{\bP}{\mathbf{P}}
\newcommand{\bh}{\bm{h}}
\newcommand{\bn}{\hat{\bm{n}}}
\newcommand{\ms}{\mathcal{S}}
\newcommand{\dd}{\mathrm{d}}
\newcommand{\norm}[1]{\Vert{#1}\Vert}
\newcommand{\pf}[2]{\frac{\partial #1}{\partial #2}}

\section{Scene Recovery from a Few ToF Pixels}
\label{sec:method}

\noindent \textbf{Problem Formulation}. 
We aim to recover 3D geometry of a \textit{Lambertian} scene specified by a set of parameters $\bP$ from a distributed set of $n$ ToF sensors with known poses. 
We make two key assumptions regarding the \textit{sensing setup} and \textit{scene representation}.
\textit{First}, we assume that each ToF sensor operates via a co-located diffuse light source with a finite field-of-view, and reports a transient histogram $\bh$ which captures the intensity of light returning from the scene after a controlled pulse of illumination. This assumption covers a range of ToF sensors, including flash LiDAR and the low-cost SPAD sensor considered in this paper.
\textit{Second}, we utilize a mesh-based 3D scene representation, where the scene is modeled as a polygonal mesh composed of interconnected triangles that define its shape and surface. Mesh-based representations are widely used in graphics and many shape models are built on meshes~\cite{MANO:SIGGRAPHASIA:2017,FLAME:SiggraphAsia2017,SMPL:2015,xu_visual-tactile_2023}. In this case, $\bP$ can be the 6 DoF pose of a 3D object mesh or parameters for a mesh-based shape model. 
Our goal is to estimate $\bP$ from the set of measured histograms $\{ \bh_i \}_{i=1}^{n}$.

\smallskip
\noindent \textbf{Method Overview}. Our method consists of two steps: 1) given a set of input transient histograms $\{ \bh_i \}_{i=1}^{n}$, a feedforward network outputs a prediction $\bP_{\text{FF}}$ of the scene parameters, and 2) an analysis-by-synthesis based refiner takes $\bP_{\text{FF}}$ as an initial estimate, alongside camera pose and any other scene prior (\eg, a parametric model), and iteratively optimizes $\bP_{\text{FF}}$ to minimize the difference between the measured histograms $\{ \bh_i \}_{i=1}^{n}$ and histograms synthesized by our differentiable renderer. An illustration of our method as applied to the task of 6D pose estimation is shown in~\cref{fig:method_diagram}. In what follows we introduce the SPAD image formation process and present each component of our method.

\subsection{Background: Transient Formation Model}
\label{subsec:sensor_model}
We utilize physics-based sensor modeling to accurately render the transient histograms $\{ \bh_i \}_{i=1}^{n}$. For each captured histogram, the laser source emits $N_{\text{emit}}$ photons. Assuming that the source is co-located with the sensor at the origin $\bm{o}$, the rays of the emitted photons can be parametrized by a direction $\bw$. As in~\cite{mu_towards_2024, avidan_3d_2022, sifferman_unlocking_2023}, we ignore high-order light paths and consider one-bounce paths only. Therefore, each photon travels from $\bm{o}$ to a point on the scene $\bx$ and then back to $\bm{o}$, where $\bx=\bx(\bw,\mathcal{M})$ is the first intersection between the ray in the direction of $\bw$ and the scene $\mathcal{M}$.

Following prior work~\cite{mu_towards_2024, avidan_3d_2022}, the expected number of photons $N[i]$ received by the sensor within the $i$-th bin, in its angular integral form, is
\begin{equation}
\resizebox{.43\textwidth}{!}{
    $N[i] = N_{\text{emit}}\int_{\Omega}\ I(\bw)\frac{\rho(\bx)}{\pi}\frac{\big\langle-\bw,\bn(\bx)\big\rangle}{\norm{\bx}^2}  W(\frac{2\norm{\bx}}{c}, t_{i})\ \dd\bw$, 
}\label{eqn:render}
\end{equation}
where $\Omega$ is the space of solid angles within the FoV of the sensor. $I(\bw)$ encodes the intensity of the laser along the direction $\bw$. $\rho(\bx)$ is the albedo, and $\bn(\bx)$ is the normal of the surface at $\bx$. $t_i=i\Delta t $ corresponds to the time of the leading edge of the $i^{\text{th}}$ bin, with $\Delta t$ as the bin width. Lastly, 
\begin{equation}\label{eqn:box}
    W(t, t_i)=\begin{cases}
    1 & \text{if}\ \  t\in[t_i,t_i+\Delta t),\\
    0 & \text{otherwise}
    \end{cases} 
\end{equation}
bins the photons. Due to hardware effects, the binning of photons is not perfect in reality. $\bh[i]$ in fact might also detect photon arrivals near but outside that bin, and is affected by the shape of the outgoing laser pulse. Therefore, we convolve the expected photon numbers with an empirically derived discrete jitter kernel $\bs$ to account for this effect
\begin{equation}
    \bh[i] = \Sigma_j N[i+\Delta i-j]\bs[j],
\end{equation}
where $\Delta i$ is sensor-specific, constant temporal offset to the histogram, which needs to be calibrated.

\smallskip
\noindent
\textbf{Pile-up Correction.} Previous works~\cite{gupta_photon-flooded_2019, mu_towards_2024} model the pile-up effect, which can significantly alter the measured histogram at high levels of ambient or active flux. To mitigate this effect, many existing sensors pre-process the histogram with algorithms like Coates' correction~\cite{coates1968correction_}. With this correction and under reasonable lighting conditions, the pile-up effect is often negligible. We thus do not include pile-up in our imaging model.




\subsection{Differentiable Rendering}
\label{subsec:renderer}

We numerically integrate \cref{eqn:render} using the weighted sum
\begin{equation}
\resizebox{.43\textwidth}{!}{
    $N[i] \approx \frac{N_{\text{emit}}}{\pi}\sum_{\bw\in\mathcal{W}}Q(\bw)I(\bw)\rho(\bx)\frac{\big\langle-\bw,\bn(\bx)\big\rangle}{\norm{\bx}^2}  W(\frac{2\norm{\bx}}{c}, t_{i})$, 
}\label{eqn:numeric}
\end{equation}
where $\mathcal{W}$ is a specified set of rays, and $Q(\bw)$ is the associated quadrature.
To make the rendering differentiable, we calculate $\partial{\bx}/\partial{\mathcal{M}}$ and $\partial{\bn}/\partial{\mathcal{M}}$ using off-the-shelf differentiable rendering libraries. Specifically, we set $\mathcal{W}$ to a $h\times w$ grid of rays, fully covering the FoV and resembling the rendering of classical pixels, and the rasterization computes the rays' $\bx, \bn$, and the gradients.
Suppose that the center of the FoV is the z-axis and the imaging plane is $z=1$, each pixel has area $A_\text{pixel}=(2\tan(FoV/2))^2 / \ (h\cdot w).$
Assuming $\bw=(\omega_x,\omega_y,\omega_z)$, the quadrature $Q(\bw)$ transforms the square pixel areas to solid angle differentials by
\begin{equation}
Q(\bw)=\frac{A_\text{pixel}\langle\omega, \hat{z}\rangle}{(1/\omega_z)^2}=\frac{4\tan^2({FoV}/{2})\omega_z^3}{h\cdot w}.
\end{equation}

The binning function $W$ in~\cref{eqn:box} is discontinuous. We thus approximate it with the sigmoid function $\sigma(x)$ by
\begin{equation}
    W(t,t_i)=\sigma\left( k(t-t_i) \right)-\sigma \left( k(t-t_i-\Delta t) \right), 
\end{equation}
where $k$ is a hand-picked constant to balance smoothness and realism. Further, the intensity map $I(\bw)$ is discontinuous under an idealized diffuse laser source, since it provides uniform illumination within its FoV and zero illumination elsewhere. However, real-world lasers exhibit a non-uniform distribution, where intensity is highest at the center and gradually decreases toward the edges of the FoV. This allows us to approximate $I(\bw)$ using a differentiable, spatially-varying function. We fit this function using real-world sensor properties in our experiments (see \cref{subsec:implementation_details}).

\subsection{Feedforward Estimation of Scene Parameters}
\label{subsec:ff_network}
We learn a neural network $f_\theta$ to predict initial scene parameters $\bP_{\text{FF}}$, which will be further refined. This is given by
\begin{equation}
    f_\theta(\{\bh_i\}_{i=1}^n) \rightarrow \bP_{\text{FF}},
\end{equation}
where $\theta$ is the network weights learned from data, $\{\bh_i\}_{i=1}^n$ is the input of $n$ transient histograms from ToF sensors. Namely, this feedforward network directly regresses the scene parameters based on sensor data. 

\smallskip
\noindent \textbf{Network Architecture.} Specifically, $f_\theta$ is a standard Transformer model~\cite{vaswani2017attention}. The input transient histograms are first normalized, and then embedded using an MLP. These embeddings are added to positional embeddings, and further processed by a stack of Transformer blocks (4 in our implementation). The output embeddings are concatenated and fed into another MLP to predict scene parameters $\bP$. This network is trained with full supervision, and the loss function varies depending on the scene parameterization used, as described in the supplement. 

\smallskip
\noindent \textbf{Sim-to-Real Transfer.} A key challenge for training is the lack of real-world sensor data. We explore training on a large-scale synthetic dataset and transfer the learned model directly to real-world captures. This is made possible thanks to our efficient renderer in~\cref{subsec:renderer} and availability of 3D models~\cite{calli_ycb_2015}. We demonstrate strong results using this sim-to-real transfer in our experiments.

\smallskip
\noindent \textbf{Discussion.} Our network assumes fixed sensor poses and requires re-training for every sensor configuration. This design is highly tailored for our current hardware prototypes, yet can be easily extended to accommodate varying sensor poses, \eg, encoding sensor pose as part of the input~\cite{li2025cameras}.

\subsection{Parameter Refinement}
Given an estimate $\bP_\text{FF}$ of the scene parameters from the feedforward network and the differentiable renderer $\mathcal{R}$, we propose an analysis-by-synthesis approach to further refine $\bP_\text{FF}$. This is done by directly optimizing $\bP$ to minimize the difference between the measured histograms $\{\bh_{i}\}_{i=1}^n$ and rendered histograms $\mathcal{R}(\bP)$, given by
\begin{equation}
    \arg\min_\mathbf{P} \ \Sigma_{i=1}^n \|\mathcal{R}(\bP)_i - \bh_{i}\|.\label{eqn:refinement}
\end{equation}
Since the renderer $\mathcal{R}$ is fully differentiable, gradient descent can be used to solve this optimization locally. Specifically, this optimization starts from the initial estimate $\bP_\text{FF}$ and applies gradients steps until convergence. Since the rendering process $\mathcal{R}$ is highly nonlinear, the quality of the solution depends on the accuracy of initial estimate $\bP_\text{FF}$.

\section{Experiments on 6D Pose Estimation}
\label{sec:experiments}
To adopt our method for 6D pose recovery, we set the scene parameters $\mathbf{P}$ to a rotation $\mathbf{R}$ and translation $\mathbf{T}$ which transform a known object mesh to its position in a global coordinate frame. We evaluate our method for 6D pose estimation in simulation and on real-world captures.

\subsection{Hardware and Real-world Capture} \label{subsec:real_dataset_capture}
\begin{figure}[t!]
    \centering
    \includegraphics[width=0.9\linewidth]{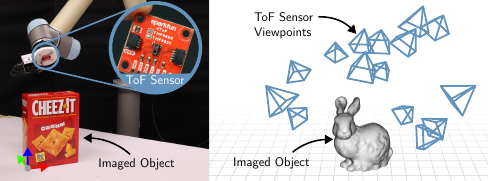}\vspace{-0.5em}
    \caption{\textbf{Left}: Illustration of our capture setup, where a ToF sensor is mounted on a robot arm and moved between a set of positions. \textbf{Right}: 15 sensor poses used for our experiments.}\vspace{-1.5em}
    \label{fig:capture_setup}
\end{figure}

\noindent\textbf{Hardware Prototype}. Our imaging system assumes known relative position of the ToF sensors. In practical deployment, this may be achieved by placing multiple sensors in a stationary position in the environment, or by attaching them each to a rigid object, like a mobile device. To allow flexibility for our experiments, we instead place a single sensor on an industrial robot arm, and move the sensor to multiple positions by controlling the robot while the scene remains static. We rely on the robot's kinematics, which are quoted as repeatable to $\pm 0.1\text{mm}$~\cite{UR5Datasheet2016}, to record sensor pose. 

We use the AMS TMF8820 SPAD-based ToF sensor.
Like other sensors of its class, the sensor is very small ($12.8\text{mm}^3$, $<1$g) and low-power ($<100\text{mW}$)~\cite{TMF8820_}. The illumination source is an integrated low-power 940nm VCSEL laser. We operate the sensor in ``low range, high accuracy mode,'' and 4 million iterations per measurement, giving it a maximum range of $1.5$m and a bin size equivalent to $\sim 1.4\text{cm}$.
We interface with the sensor via an attached microcontroller, which forwards transient histogram measurements from the sensor to a connected computer. \smallskip

\noindent \textbf{Capture Setup.} We sample random sensor positions between 30cm and 80cm from the workspace center (within the range of TMF8820), and random orientations which face the camera to the workspace center ($\pm 15^\circ$). For a fair comparison, the same 15 randomly sampled sensor poses are used for all real-world experiments. This number (15) allows for practical data capture, and was chosen to strike a balance between estimation accuracy and information sparsity based on our simulation results (see \cref{subsec:main_results}). An Intel RealSense D405 depth-from-stereo camera is affixed next to the ToF sensor for ground truth capture. We utilize the merged point cloud from the depth camera views alongside ICP~\cite{besl_method_1992} and manual registration to generate ground truth object poses. We measure the geometry of the known background (the tabletop) by touching the robot to the surface at multiple points. This capture setup is shown in~\cref{fig:capture_setup}.\smallskip

\noindent \textbf{Data Capture.} We capture each object at 25 poses (10 for the highly symmetric basketball, softball, and tennis ball) by manually positioning the object such that the object center is within 15cm of the workspace center. Effort is made to distribute the object poses uniformly within the workspace and to vary the object orientation uniformly. Because the objects are placed on a tabletop, we are restricted to orientations that provide stable support on the surface.

\subsection{Implementation Details} \label{subsec:implementation_details}
\textbf{TMF8820 Modeling.}
The TMF8820 reports transient histograms for nine separate fields-of-view, called ``zones'', each of which correspond to different sets of pixels on the SPAD array. Significant bloom artifacts are present between zones, and the exact zone dimensions are poorly defined~\cite{sifferman_unlocking_2023, behari_blurred_2024}. To address this challenge, and in line with our focus on very-low-pixel-count regimes, we aggregate the zones into one by summing across the zone dimension, yielding a single 128-bin histogram per sensor measurement. This approach is consistent with prior work~\cite{mu_towards_2024, sifferman_using_2024}. 

Further, in the TMF8820 datasheet, the laser illumination is reported as non-uniform across the FoV~\cite{TMF8820_} $I(\bw)$. We therefore approximate the intensity map using
\begin{equation}
\label{eqn:laser_intensity}
I(\bw)=K_1\exp\left(-K_2\big(\omega_x^2+\omega_y^2\big)-K_3\big(\omega_x^4+\omega_y^4\big)\right).
\end{equation}
We calibrate the constants $K$ to match the intensity map shown in the TMF8820 datasheet. A visualization of $I(\bw)$ is included in the supplement (\cref{fig:laser_intensity}).
We find that pile-up is not very apparent even at high ambient light levels. 

\smallskip
\noindent
\textbf{Jitter Kernel.}
The TMF8820 reports the shape of the outgoing laser pulse in a ``reference histogram'' alongside each measurement, which is measured from a SPAD sensor inside the laser cavity. Because this reference histogram is itself captured by a SPAD sensor, it encapsulates the shape of the outgoing laser pulse and the temporal response function of the SPAD itself. We make use of this histogram as the jitter kernel $\bs$ in our imaging model.

\smallskip
\noindent
\textbf{Sensor Calibration.}
The reference histogram reported by the TMF8820 is not reported at the same temporal resolution as the transient histogram~\cite{sifferman_unlocking_2023, mu_towards_2024}. We resample the reference histogram by a factor of $\bs_\text{scale}$ before use in our imaging model. Additionally, the temporal resolution $\Delta t$ and a constant temporal offset to the histogram $\Delta i$ are not known. Following prior work, we recover the parameters $\bs_\text{scale}$, $\Delta t$, and $\Delta i$ by calibrating on some set of reference captures of a planar surface with known geometry. To do so, we keep scene geometry fixed, and optimize the sensor parameters to minimize the loss between captured and rendered histograms, akin to~\cref{eqn:refinement}.

\begin{figure}[t!]
    \centering
    \includegraphics[width=0.9\linewidth]{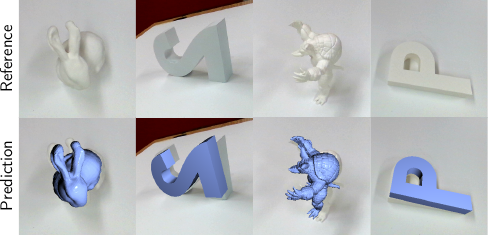}\vspace{-0.5em}
    \caption{\textbf{Visualization} of 6D pose estimation results with 3D-printed objects using our method (feedforward + refiner). For each object, the pose prediction with the median pose error (ADD) over the 25-capture dataset is shown.}\label{fig:6d_pose_3d_printed_objs}\vspace{-1.5em}
\end{figure}

\smallskip
\noindent
\textbf{Feedforward Network.}
We train the network described in~\cref{subsec:ff_network} to predict 6D pose. For non-symmetric objects, we use a combination of rotation loss, translation loss, and a point matching loss. For symmetric objects, we use the ADD-S loss~\cite{xiang_posecnn_2018}. See the supplement (\cref{sec:6d_pose_estimation_loss_fn}) for details of the loss functions and training parameters.
A single-instance forward pass takes $\sim4.6$ms 
on Nvidia RTX 4080.

\smallskip
\noindent
\textbf{Pose Refinement.}
We use adaptive gradient descent following Adam~\cite{kingma_adam_2017} to solve~\cref{eqn:refinement}. We set the step size (\ie, learning rate) to 0.01 for $\mathbf{R}$ and 0.001 for $\mathbf{T}$. Additionally, we optimize the albedo of both the object ($\rho_\text{obj}$) and the planar surface ($\rho_\text{plane}$). These albedos are not predicted by the feedforward network; instead they are initialized to 1 and further optimized. We find empirically that optimizing albedos improves performance. We set the number of optimization steps to 200. We represent $\mathbf{R}$ using the 6D rotation representation proposed in~\cite{zhou_continuity_2019}. Differentiable rendering is implemented via Nvdiffrast~\cite{Laine2020diffrast} and PyTorch~\cite{pytorch}.

\subsection{Experiment Protocol} \label{subsec:experiment_protocol}

\noindent
\textbf{Datasets.} We evaluate our method for 6D pose estimation on two sets of objects: 1) 3D printed test objects and 2) seven readily available objects from the YCB dataset~\cite{calli_ycb_2015} --- a standard benchmark for 6D pose recognition. See the supplement (\cref{fig:objects}) for images of the objects. For YCB objects, the high-resolution ``Google 16k'' meshes provided by the YCB dataset are used for data generation and as input to the refiner. A child's basketball is used in place of the child's soccer ball in the YCB object set, along with a manually constructed spherical mesh.\smallskip

\noindent
\textbf{Synthetic Training Data.}
We generate synthetic data with our renderer to train the feedforward model. For each object, we synthesize 200K samples by limiting the object center within $15$cm of the workspace center. Object orientations are randomly sampled. To ensure physical plausibility, the object height is adjusted so that at least one vertex of the mesh lies on the planar surface, preventing the object from appearing to float in space, though this configuration may not correspond to a stable resting pose. This setup imposes a conservative prior on object placement. The planar surface is included in the scene when rendering synthetic data. We also perform domain randomization~\cite{tobin2017domain}; we add Gaussian noise with a standard deviation of $1.5$cm to the sensor positions independently for each sample, and vary the albedo of the object and the planar surface.

\begin{table*}[th!]
\centering
{
\footnotesize
\begin{tabular}{lllcccccccc}
\toprule
& \multirow{2}{*}{\makecell[l]{\textbf{Total}\\\textbf{Pixels}}} & & \multicolumn{8}{c}{\textbf{AUC-ADD-S} (\textuparrow)} \\
\textbf{Data} & & \textbf{Method} & \textbf{Crackers} & \textbf{Mustard} & \textbf{Chips} & \textbf{SPAM} & \textbf{Basketball} & \textbf{Tennis Ball} & \textbf{Softball} & \textbf{Mean} \\
\midrule
Sim. & 15 & 1 Px Point Cloud$^\dagger$ & 78.36   & 82.12   & 77.83   & 85.07          & 82.92   & 88.09            & 86.36            & 82.96 \\
Real & 15 & Ours: Feedforward    & 88.02   & 90.04   & 88.38   & \textbf{90.04} & 95.15   & \textbf{96.26}   & \textbf{94.95}   & 91.83 \\
Real & 15 & Ours: FF + Refiner       & \textbf{90.04}   & \textbf{90.07}   & \textbf{88.50}   & 90.00 & \textbf{95.76}   & 96.06    & \textbf{94.95}   & \textbf{92.20} \\
\midrule
\graytext{Sim}. & \graytext{3840} & \graytext{$16^2$ Point Cloud$^\dagger$} & \graytext{95.17} & \graytext{97.23} & \graytext{97.23} & \graytext{97.19} & \graytext{97.67} & \graytext{97.57} & \graytext{97.37} & \graytext{97.06} \\
\graytext{Real} & \graytext{407K} & \graytext{Single-View RGB~\cite{leonardis_foundpose_2025}} & \graytext{60.71} & \graytext{87.93} & \graytext{40.12} & \graytext{58.95} & \graytext{65.46} & \graytext{77.68} & \graytext{72.42} & \graytext{66.18} \\
\graytext{Real} & \graytext{407K} & \graytext{Single-View RGB-D$^*$~\cite{wen_foundationpose_2024}} & \graytext{90.49} & \graytext{92.10} & \graytext{92.54} & \graytext{93.80} & \graytext{94.24} & \graytext{86.67} & \graytext{94.24} & \graytext{92.01} \\
\bottomrule
\end{tabular}
}\vspace{-0.5em}
\caption{\textbf{Results of 6D pose estimation with (symmetric) objects from the YCB object set}. \graytext{gray}: methods using additional pixels; \\ $^\dagger$: methods using oracle ground-truth pose; $^*$: methods using simulated high-resolution point cloud data. See details in \cref{subsec:experiment_protocol}.}\vspace{-0.8em}
\label{tab:6d_pose_results_ycb}
\end{table*}

\begin{table*}[th!]
\centering
{ \footnotesize
\begin{tabular}{lllcccccc}
\toprule
& \multirow{2}{*}{\makecell[l]{\textbf{Total}\\\textbf{Pixels}}} & & \multicolumn{6}{c}{\textbf{AUC-ADD} (\textuparrow)} \\
\textbf{Data} & & \textbf{Method} & \textbf{Two} & \textbf{P} & \textbf{L} & \textbf{Bunny} & \textbf{Armadillo} & \textbf{Mean} \\
\midrule
Sim. & 15 & 1 Px Point Cloud$^\dagger$     & 73.87   & 59.65   & 55.11   & 56.14   & 53.89  & 59.73 \\
Real & 15 & Ours: Feedforward     & 74.67   & 77.31   & 69.11   & 67.78   & 65.93  & 70.96 \\
Real & 15 & Ours: FF + Refiner        & \textbf{83.47}   & \textbf{84.94}   & \textbf{77.75}   & \textbf{77.68}   & \textbf{79.71}  & \textbf{80.71} \\
\midrule
\graytext{Sim.} & \graytext{3840} & \graytext{$16^2$ Point Cloud$^\dagger$} & \graytext{97.55} & \graytext{96.18} & \graytext{95.17} & \graytext{96.18} & \graytext{96.90} & \graytext{96.39} \\
\graytext{Real} & \graytext{407K} & \graytext{Single-View RGB~\cite{leonardis_foundpose_2025}} & \graytext{56.99} & \graytext{65.40} & \graytext{51.51} & \graytext{64.63} & \graytext{86.28} & \graytext{64.96} \\
\graytext{Real} & \graytext{407K} & \graytext{Single-View RGB-D$^*$~\cite{wen_foundationpose_2024}} & \graytext{86.57} & \graytext{85.51} & \graytext{82.72} & \graytext{88.30} & \graytext{87.58} & \graytext{86.14} \\
\bottomrule
\end{tabular}
}\vspace{-0.5em}
\caption{\textbf{Results of 6D pose estimation with (non-symmetric) 3D printed objects}. \graytext{gray}: methods using additional pixels; $^\dagger$: methods using oracle ground-truth pose; $^*$: methods using simulated high-resolution point cloud data. See details in \cref{subsec:experiment_protocol}.}
\label{tab:6d_pose_results_3dp}\vspace{-1.5em}
\end{table*}

\begin{figure}[t!]
    \centering
    \includegraphics[width=0.9\linewidth]{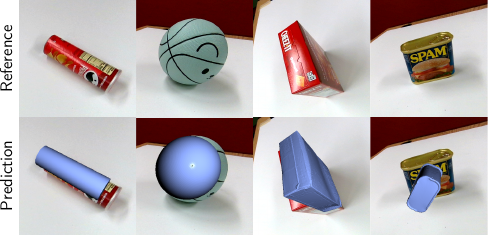}\vspace{-0.5em}
    \caption{\textbf{Visualization} of 6D pose estimation results with objects from the YCB dataset using our method (feedforward + refiner). For each object, the pose prediction with the median pose error (ADD-S) over the 25-capture dataset is shown. 
    }\vspace{-1.5em} \label{fig:6d_pose_ycb_objs}
\end{figure}

\smallskip
\noindent
\textbf{Evaluation Metrics.} We follow standard metrics~\cite{xiang_posecnn_2018} for evaluation, including ADD for non-symmetric objects, and ADD-S for symmetric objects. ADD captures the average distance between corresponding points on the predicted and ground truth object. ADD-S captures the average distance between a point on the predicted object and \textit{the nearest} point on the ground truth object. We report the AUC-ADD(-S) in order to capture the distribution of scores over the entire dataset, with the maximum threshold in calculating AUC set to 10cm.
Note that ADD(-S) score is highly dependent on the scale and geometry of a specific object, thus should not be compared between objects.

\smallskip
\noindent
\textbf{Point Cloud Baselines.}
We implement a point-cloud based baseline which represents an upper bound on point-cloud based system performance. To avoid pitfalls of any one particular depth camera, we simulate \textit{idealized} sensor measurements; for each sensor pixel we project rays to get points of intersection with scene geometry, which are combined to form a point cloud. Points which lie on the planar surface are removed. The result is a noise-free point cloud of points on the object mesh. We then use ICP~\cite{besl_method_1992} to align the object mesh to this point cloud. To maximize the chance of success, the ground truth pose is used as initialization for ICP. 
Therefore, this baseline represents the best possible performance of a vanilla ICP-based registration system.
%
We consider two variants: single-pixel, in which the same number of imaging pixels are used as in our system (one per view), and $16^2$, in which a $16\times16$ point cloud sampled from a pixel grid spanning an FoV matching the diffuse sensor is generated \textit{per view} (yielding $256\times$ imaging pixels of our system). 


\smallskip
\noindent
\textbf{RGB(D) Baselines.}
For reference, we compare to two recent deep models: FoundPose~\cite{leonardis_foundpose_2025}, which uses single-view RGB input, and FoundationPose~\cite{wen_foundationpose_2024}, which uses single-view RGB-D input. For both methods, the RGB image from viewpoint 15 is used as it provides a wide view. For FoundationPose, a high resolution depth camera view is simulated from the ground truth object pose.

\subsection{6D Pose Estimation Results}\label{subsec:main_results}

\noindent \textbf{Main Results}. Our main results are presented in~\cref{tab:6d_pose_results_ycb} and~\cref{tab:6d_pose_results_3dp}. Visualization of sample results with median pose errors are shown in~\cref{fig:6d_pose_3d_printed_objs} and~\cref{fig:6d_pose_ycb_objs}. In all cases, our method outperforms the best-case performance of using a single-pixel point cloud. Additionally, our refiner improves performance in most cases. We also notice that our refiner yields a much larger performance improvement over the feedforward network on 3D printed objects. We hypothesize that this is due to the ambiguity of symmetric objects, and the symmetric loss can easily converge to a local minimum. Despite reasonable metrics, the SPAM is a failure case for our method; because the object is very small and near-symmetric along many axes, a high ADD-S score can be achieved by matching object translation. However, qualitative results (\cref{fig:6d_pose_ycb_objs}) show that orientation is not predicted reliably. Our method approaches the performance of the RGB-D baseline, while exceeding the performance of the RGB baseline which struggles due to the lack of metric depth information and/or lack of object texture.


\smallskip
\noindent \textbf{Varying Viewpoint Count}.
In simulation, we evaluate the performance of our method with varying number of views. We compare to the point cloud-based baseline described in~\cref{subsec:experiment_protocol}. We evaluate on synthetic data of the 25 real ``2'' poses. Camera poses are sampled via the same process as real poses (~\cref{subsec:real_dataset_capture}). Results are shown in the rightmost panel of~\cref{fig:teaser}. See the supplement for full results (\cref{tab:6d_pose_views}). Our approach exceeds the performance of the point cloud-based baseline for $5$-$100$ total pixels. The point cloud-based method suffers with very few input pixels because, despite some variation in sensor orientation, the poses do not achieve good coverage of the scene.\smallskip

\noindent\textbf{Additional Results}. Due to space limit, the following are included in the supplement: (1) results with varying scene reflectance and ambient light; (2) ablations on the sensor model and feedforward network training; and (3) experiments on sensor interference. 

\section{Exploration Beyond 6D Pose Estimation}
\label{sec:extensions}

\subsection{Size and Position of Spherical Objects}
\label{subsec:extension_spheres}
We experiment with recovering the size and location of a sphere resting on a planar surface
We use an identical method to that used for 6D pose estimation (\cref{sec:experiments}), except the predicted parameters $\mathbf{P}$ consist of the center point and diameter of a sphere, rather than the rotation and translation of a known mesh. Both parameters are predicted by the feedforward network and optimized during refinement. We evaluate on our pre-existing captures of the basketball, softball, and tennis ball objects. The results of this experiment are shown in~\cref{tab:sphere_recovery}. We find that our method can recover the position and diameter of the sphere with an average error of $<1$cm in all cases, with often $<0.5$cm of error, despite the temporal resolution of an individual SPAD being $\sim1.4$cm.

\subsection{Human Hand Pose}
\label{subsec:extension_hand}
\begin{figure}[t!]
    \centering
    \includegraphics[width=0.85\linewidth]{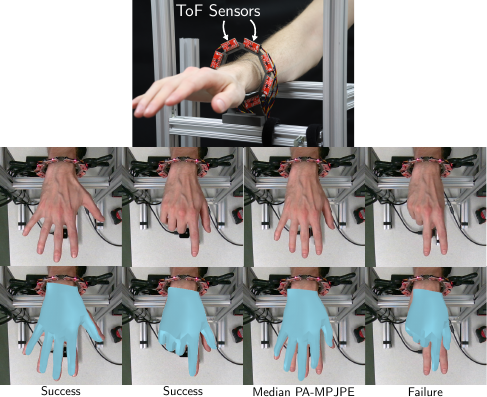}\vspace{-0.5em}
    \caption{\textbf{Top}: Setting for hand pose capture, in which eight ToF sensors encircle the wrist. \textbf{Bottom}: Visualization of results of applying our method to hand pose estimation, corresponding to method "Pretrained on Sim., F.T. on Real" in~\cref{tab:hand_recovery}.}
    \label{fig:hand_recovery}\vspace{-1.5em}
\end{figure}

We recover human hand pose (absolute pose and articulation) from a ring of eight sensors encircling the wrist, as shown in~\cref{fig:hand_recovery}. We modify the feedforward prediction to include pose, shape, global translation, and rotation of the human hand. The hand pose and shape is represented using the parameters of the MANO hand model~\cite{MANO:SIGGRAPHASIA:2017}.\smallskip

\begin{table}[th!]
    \centering
    { \footnotesize
    \begin{tabular}{lccc}
    \toprule
    \textbf{Object} & \makecell{\textbf{Mean Error}\\ \textbf{in Diameter (cm)} (\textdownarrow)} & \makecell{\textbf{Mean Error}\\ \textbf{in Position (cm)} (\textdownarrow)} \\
    \midrule
    Basketball & 0.35 (1.9\%) & 0.84 \\
    Softball & 0.28 (2.9\%) & 0.24 \\
    Tennis Ball & 0.33 (4.8\%) & 0.39 \\
    \bottomrule
    \end{tabular}
    }\vspace{-0.5em}
    \caption{\textbf{Results} of recovering position / size of \textbf{spherical objects}.}
    \label{tab:sphere_recovery}\vspace{-0.8em}
\end{table}

\begin{table}[th!]
    \centering
    { \footnotesize
    \begin{tabular}{lcc}
    \toprule
    \textbf{Method} & \textbf{PA-MPJPE (mm)} (\textdownarrow) \\
    \midrule
    ToF-based Prior Work$^\dagger$~\cite{devrio_discoband_2022} & 11.96 \\
    Trained on Sim. Data Only (ours) & 19.56 \\
    Trained on Real Data Only (ours)& 9.98 \\
    P.T. on Sim., F.T. on Real (ours)& 8.18 \\
    RGB-Based Method$^\dagger$~\cite{pavlakos_reconstructing_2024} & 6.0 \\
    \bottomrule
    \end{tabular}
    }\vspace{-0.5em}
    \caption{\textbf{Results of hand pose and shape estimation}. $^\dagger$Related works are provided for context only; metrics are over a different dataset and should not be directly compared.}
    \label{tab:hand_recovery}\vspace{-1em}
\end{table}

\noindent \textbf{Data Capture.} As an initial feasibility study, we gather 250 measurements of a single individual's hand from the ring of sensors. To sample hand poses, we prompt the user to match their hand pose to a random hand pose selected from the DART dataset~\cite{gao_dart_2022}. RGB-D cameras are mounted above and below the hand to capture ground truth, which is provided by the RGB-based method HaMeR~\cite{pavlakos_reconstructing_2024}, and aligned to a fused point cloud from the two depth maps via ICP~\cite{besl_method_1992}. We reserve 50 captures for testing.\smallskip

\noindent \textbf{Results.} We report Procrutes aligned mean per joint position error (PA-MPJPE), a standard metric for hand tracking~\cite{devrio_discoband_2022, pavlakos_reconstructing_2024}. PA-MPJPE captures the average distance between corresponding joints in the predicted and ground truth hand pose. The results of our experiment are shown in~\cref{tab:hand_recovery}.
We find that, in this setting, training on simulated data alone yields unsatisfactory results. A closer inspection reveals that the simulated histograms become inaccurate at distances below 15cm. 
We attribute this to unmodeled sensor effects, such as unmodeled effects such as gating and/or pile-up from the high intensity of returning light.
While we attempted to mitigate this issue by modeling these effects and learning a custom gating function, these efforts did not lead to improved performance. For the same reason, our refiner module is also not effective for hand pose estimation.


We observe significantly improved results when training on real data, with the best results achieved through a two-stage process: pre-training on simulated data followed by fine-tuning on real data. 
Despite the limited realism of the simulated data, pre-training still provides benefits, likely because the network is able to learn transferable high-level features.
To contextualize our results, we include comparisons to related works in~\cref{tab:hand_recovery}.
However, results from prior works are based on different datasets and experimental conditions, and thus are not directly comparable.

\section{Discussion}
\label{sec:discussion}

Our work demonstrate that a few (\eg, $15$) diffuse ToF pixels are sufficient to recover simple scene geometry. Moreover, we showcased the potential of our approach for more complex geometry, including recovering the position and scale of a spherical object, and human hand pose estimation. Our work offers an initial step toward enabling a range of practical applications and open up several promising directions for future research.


\smallskip
\noindent \textbf{Practical Implications.} 
While still in its early stage, our approach has great potential for 3D vision applications that benefit from low-cost, low-power, and distributed sensing. One promising application domain is \textit{wearable computing}. Inspired by our experiments on hand tracking, we envision that an array of miniature ToF sensors could be deployed in head-mounted or wrist-worn devices to track a user's body motion (\eg, arm and hand pose), enabling gesture-based user interfaces. Another key application domain is \textit{robotics}. Imagine a robotic arm or drone equipped with a distributed array of lightweight, energy-efficient ToF sensors. These sensors could function as a network of spatially distributed cameras, reconstructing the environment from an inside-out perspective, enhancing tasks like grasping, navigation, and human-robot interaction. 

\smallskip
\noindent \textbf{Future Directions.} 
Our work demonstrates the estimation of 6D pose of rigid objects in a tabletop setting. Future work should aim to improve robustness to environmental factors such as ambient light and varied surface reflectance. This could be achieved by explicitly modeling these factors or developing methods that are inherently invariant to them. Additionally, future work should explore recovery of more complex scene geometries at larger scales, \eg, multiple deformable, articulated objects in room- or playground-sized environments. A promising future direction is learning from large-scale synthetic data. Encouragingly, our results have demonstrated that effective sim-to-real transfer is possible with ToF histogram data. We are hopeful that large-scale synthetic data can be applied to a range of inference tasks.

\medskip
\noindent{\textbf{Acknowledgements:} This work was was supported in part by the National Science Foundation under Grant Numbers 2152163 (NRT), 2333491 (CPS Frontier), 2442739 (CAREER), 1943149 (CAREER), 2107060 (CNS Core), by ARL under contract number W911NF-2020221, and by ONR grant N000142412155.
}

{
    \bibliographystyle{ieeenat_fullname}
    \bibliography{main}

\begin{thebibliography}{52}
\providecommand{\natexlab}[1]{#1}
\providecommand{\url}[1]{\texttt{#1}}
\expandafter\ifx\csname urlstyle\endcsname\relax
  \providecommand{\doi}[1]{doi: #1}\else
  \providecommand{\doi}{doi: \begingroup \urlstyle{rm}\Url}\fi

\bibitem[AG()]{TMF8820_}
AMS~OSRAM AG.
\newblock \emph{TMF882X Datasheet}.
\newblock AMS OSRAM AG.

\bibitem[Ansel et~al.(2024)Ansel, Yang, He, Gimelshein, Jain, Voznesensky, Bao, Bell, Berard, Burovski, Chauhan, Chourdia, Constable, Desmaison, DeVito, Ellison, Feng, Gong, Gschwind, Hirsh, Huang, Kalambarkar, Kirsch, Lazos, Lezcano, Liang, Liang, Lu, Luk, Maher, Pan, Puhrsch, Reso, Saroufim, Siraichi, Suk, Zhang, Suo, Tillet, Zhao, Wang, Zhou, Zou, Wang, Mathews, Wen, Chanan, Wu, and Chintala]{pytorch}
Jason Ansel, Edward Yang, Horace He, Natalia Gimelshein, Animesh Jain, Michael Voznesensky, Bin Bao, Peter Bell, David Berard, Evgeni Burovski, Geeta Chauhan, Anjali Chourdia, Will Constable, Alban Desmaison, Zachary DeVito, Elias Ellison, Will Feng, Jiong Gong, Michael Gschwind, Brian Hirsh, Sherlock Huang, Kshiteej Kalambarkar, Laurent Kirsch, Michael Lazos, Mario Lezcano, Yanbo Liang, Jason Liang, Yinghai Lu, C.~K. Luk, Bert Maher, Yunjie Pan, Christian Puhrsch, Matthias Reso, Mark Saroufim, Marcos~Yukio Siraichi, Helen Suk, Shunting Zhang, Michael Suo, Phil Tillet, Xu Zhao, Eikan Wang, Keren Zhou, Richard Zou, Xiaodong Wang, Ajit Mathews, William Wen, Gregory Chanan, Peng Wu, and Soumith Chintala.
\newblock Pytorch 2: Faster machine learning through dynamic python bytecode transformation and graph compilation.
\newblock In \emph{Proceedings of the 29th ACM International Conference on Architectural Support for Programming Languages and Operating Systems, Volume 2}, page 929–947, New York, NY, USA, 2024. Association for Computing Machinery.

\bibitem[Behari et~al.(2024)Behari, Young, Somasundaram, Klinghoffer, Dave, and Raskar]{behari_blurred_2024}
Nikhil Behari, Aaron Young, Siddharth Somasundaram, Tzofi Klinghoffer, Akshat Dave, and Ramesh Raskar.
\newblock Blurred {LiDAR} for {Sharper} {3D}: {Robust} {Handheld} {3D} {Scanning} with {Diffuse} {LiDAR} and {RGB}, 2024.
\newblock arXiv:2411.19474 [eess].

\bibitem[Besl and McKay(1992)]{besl_method_1992}
P.J. Besl and Neil~D. McKay.
\newblock A method for registration of 3-{D} shapes.
\newblock \emph{IEEE Transactions on Pattern Analysis and Machine Intelligence}, 14\penalty0 (2):\penalty0 239--256, 1992.
\newblock Conference Name: IEEE Transactions on Pattern Analysis and Machine Intelligence.

\bibitem[Cai et~al.(2022)Cai, Heikkia, and Rahtu]{cai_ove6d_2022}
Dingding Cai, Janne Heikkia, and Esa Rahtu.
\newblock {OVE6D}: {Object} {Viewpoint} {Encoding} for {Depth}-based {6D} {Object} {Pose} {Estimation}.
\newblock In \emph{2022 {IEEE}/{CVF} {Conference} on {Computer} {Vision} and {Pattern} {Recognition} ({CVPR})}, pages 6793--6803, New Orleans, LA, USA, 2022. IEEE.

\bibitem[Callenberg et~al.(2021)Callenberg, Shi, Heide, and Hullin]{callenberg_low-cost_2021}
Clara Callenberg, Zheng Shi, Felix Heide, and Matthias~B. Hullin.
\newblock Low-cost {SPAD} sensing for non-line-of-sight tracking, material classification and depth imaging.
\newblock \emph{ACM Transactions on Graphics}, 40\penalty0 (4):\penalty0 1--12, 2021.

\bibitem[Calli et~al.(2015)Calli, Singh, Walsman, Srinivasa, Abbeel, and Dollar]{calli_ycb_2015}
Berk Calli, Arjun Singh, Aaron Walsman, Siddhartha Srinivasa, Pieter Abbeel, and Aaron~M. Dollar.
\newblock The {YCB} object and {Model} set: {Towards} common benchmarks for manipulation research.
\newblock In \emph{2015 {International} {Conference} on {Advanced} {Robotics} ({ICAR})}, pages 510--517, 2015.

\bibitem[Coates(1968)]{coates1968correction_}
PB Coates.
\newblock The correction for photonpile-up'in the measurement of radiative lifetimes.
\newblock \emph{Journal of Physics E: Scientific Instruments}, 1968.

\bibitem[{Dan O'shea}({2025})]{fierce2025meta}
{Dan O'shea}.
\newblock {UK start-up Singular Photonics touts SPAD sensors, Meta collab}, {2025}.
\newblock Accessed: 2025-03-07.

\bibitem[Devrio and Harrison(2022)]{devrio_discoband_2022}
Nathan Devrio and Chris Harrison.
\newblock {DiscoBand}: {Multiview} {Depth}-{Sensing} {Smartwatch} {Strap} for {Hand}, {Body} and {Environment} {Tracking}.
\newblock In \emph{Proceedings of the 35th {Annual} {ACM} {Symposium} on {User} {Interface} {Software} and {Technology}}, pages 1--13, New York, NY, USA, 2022. Association for Computing Machinery.

\bibitem[Faccio et~al.(2020)Faccio, Velten, and Wetzstein]{faccio2020non}
Daniele Faccio, Andreas Velten, and Gordon Wetzstein.
\newblock Non-line-of-sight imaging.
\newblock \emph{Nature Reviews Physics}, 2\penalty0 (6):\penalty0 318--327, 2020.

\bibitem[Gao et~al.(2022)Gao, Xiu, Li, Yang, Wang, Zhang, Zhang, Lu, and Tan]{gao_dart_2022}
Daiheng Gao, Yuliang Xiu, Kailin Li, Lixin Yang, Feng Wang, Peng Zhang, Bang Zhang, Cewu Lu, and Ping Tan.
\newblock {DART}: {Articulated} {Hand} {Model} with {Diverse} {Accessories} and {Rich} {Textures}, 2022.
\newblock arXiv:2210.07650 [cs].

\bibitem[Gupta et~al.(2019)Gupta, Ingle, Velten, and Gupta]{gupta_photon-flooded_2019}
Anant Gupta, Atul Ingle, Andreas Velten, and Mohit Gupta.
\newblock Photon-{Flooded} {Single}-{Photon} {3D} {Cameras}.
\newblock In \emph{2019 {IEEE}/{CVF} {Conference} on {Computer} {Vision} and {Pattern} {Recognition} ({CVPR})}, pages 6763--6772, Long Beach, CA, USA, 2019. IEEE.

\bibitem[Heide et~al.(2019)Heide, O’Toole, Zang, Lindell, Diamond, and Wetzstein]{heide2019non}
Felix Heide, Matthew O’Toole, Kai Zang, David~B Lindell, Steven Diamond, and Gordon Wetzstein.
\newblock Non-line-of-sight imaging with partial occluders and surface normals.
\newblock \emph{ACM Transactions on Graphics (ToG)}, 38\penalty0 (3):\penalty0 1--10, 2019.

\bibitem[Jungerman et~al.(2022)Jungerman, Ingle, Li, and Gupta]{avidan_3d_2022}
Sacha Jungerman, Atul Ingle, Yin Li, and Mohit Gupta.
\newblock {3D} {Scene} {Inference} from {Transient} {Histograms}.
\newblock In \emph{Computer {Vision} – {ECCV} 2022}, pages 401--417. Springer Nature Switzerland, Cham, 2022.
\newblock Series Title: Lecture Notes in Computer Science.

\bibitem[Kingma and Ba(2017)]{kingma_adam_2017}
Diederik~P. Kingma and Jimmy Ba.
\newblock Adam: {A} {Method} for {Stochastic} {Optimization}, 2017.
\newblock arXiv:1412.6980 [cs].

\bibitem[Kirmani et~al.(2009)Kirmani, Hutchison, Davis, and Raskar]{kirmani2009looking}
Ahmed Kirmani, Tyler Hutchison, James Davis, and Ramesh Raskar.
\newblock Looking around the corner using transient imaging.
\newblock In \emph{2009 IEEE 12th International Conference on Computer Vision}, pages 159--166. IEEE, 2009.

\bibitem[Klinghoffer et~al.(2024)Klinghoffer, Xiang, Somasundaram, Fan, Richardt, Raskar, and Ranjan]{klinghoffer2024platonerf}
Tzofi Klinghoffer, Xiaoyu Xiang, Siddharth Somasundaram, Yuchen Fan, Christian Richardt, Ramesh Raskar, and Rakesh Ranjan.
\newblock Platonerf: 3d reconstruction in plato's cave via single-view two-bounce lidar.
\newblock In \emph{Proceedings of the IEEE/CVF Conference on Computer Vision and Pattern Recognition}, pages 14565--14574, 2024.

\bibitem[Krull et~al.(2015)Krull, Brachmann, Michel, Yang, Gumhold, and Rother]{krull_learning_2015}
Alexander Krull, Eric Brachmann, Frank Michel, Michael~Ying Yang, Stefan Gumhold, and Carsten Rother.
\newblock Learning {Analysis}-by-{Synthesis} for {6D} {Pose} {Estimation} in {RGB}-{D} {Images}.
\newblock In \emph{2015 {IEEE} {International} {Conference} on {Computer} {Vision} ({ICCV})}, pages 954--962, Santiago, Chile, 2015. IEEE.

\bibitem[Labbé et~al.(2022)Labbé, Manuelli, Mousavian, Tyree, Birchfield, Tremblay, Carpentier, Aubry, Fox, and Sivic]{labbe_megapose_2022}
Yann Labbé, Lucas Manuelli, Arsalan Mousavian, Stephen Tyree, Stan Birchfield, Jonathan Tremblay, Justin Carpentier, Mathieu Aubry, Dieter Fox, and Josef Sivic.
\newblock {MegaPose}: {6D} {Pose} {Estimation} of {Novel} {Objects} via {Render} \& {Compare}, 2022.
\newblock arXiv:2212.06870 [cs].

\bibitem[Laine et~al.(2020)Laine, Hellsten, Karras, Seol, Lehtinen, and Aila]{Laine2020diffrast}
Samuli Laine, Janne Hellsten, Tero Karras, Yeongho Seol, Jaakko Lehtinen, and Timo Aila.
\newblock Modular primitives for high-performance differentiable rendering.
\newblock \emph{ACM Transactions on Graphics}, 39\penalty0 (6), 2020.

\bibitem[Li et~al.(2025)Li, Yi, Liu, Gao, Ma, and Kanazawa]{li2025cameras}
Ruilong Li, Brent Yi, Junchen Liu, Hang Gao, Yi Ma, and Angjoo Kanazawa.
\newblock Cameras as relative positional encoding.
\newblock \emph{arXiv preprint arXiv:2507.10496}, 2025.

\bibitem[Li et~al.(2017)Li, Bolkart, Black, Li, and Romero]{FLAME:SiggraphAsia2017}
Tianye Li, Timo Bolkart, Michael.~J. Black, Hao Li, and Javier Romero.
\newblock Learning a model of facial shape and expression from {4D} scans.
\newblock \emph{ACM Transactions on Graphics, (Proc. SIGGRAPH Asia)}, 36\penalty0 (6):\penalty0 194:1--194:17, 2017.

\bibitem[Li et~al.(2022)Li, Liu, Dong, Zhou, Bao, Zhang, Zhang, and Cui]{avidan_deltar_2022}
Yijin Li, Xinyang Liu, Wenqi Dong, Han Zhou, Hujun Bao, Guofeng Zhang, Yinda Zhang, and Zhaopeng Cui.
\newblock {DELTAR}: {Depth} {Estimation} from a {Light}-{Weight} {ToF} {Sensor} and {RGB} {Image}.
\newblock In \emph{Computer {Vision} – {ECCV} 2022}, pages 619--636. Springer Nature Switzerland, Cham, 2022.
\newblock Series Title: Lecture Notes in Computer Science.

\bibitem[Liu et~al.(2023)Liu, Li, Teng, Bao, Zhang, Zhang, and Cui]{liu2023multi}
Xinyang Liu, Yijin Li, Yanbin Teng, Hujun Bao, Guofeng Zhang, Yinda Zhang, and Zhaopeng Cui.
\newblock Multi-modal neural radiance field for monocular dense slam with a light-weight tof sensor.
\newblock In \emph{Proceedings of the IEEE/CVF International Conference on Computer Vision}, pages 1--11, 2023.

\bibitem[Loper et~al.(2015)Loper, Mahmood, Romero, Pons-Moll, and Black]{SMPL:2015}
Matthew Loper, Naureen Mahmood, Javier Romero, Gerard Pons-Moll, and Michael~J. Black.
\newblock {SMPL}: A skinned multi-person linear model.
\newblock \emph{ACM Trans. Graphics (Proc. SIGGRAPH Asia)}, 34\penalty0 (6):\penalty0 248:1--248:16, 2015.

\bibitem[Luo et~al.(2024)Luo, Malik, and Lindell]{luo_transientangelo_2024}
Weihan Luo, Anagh Malik, and David~B. Lindell.
\newblock Transientangelo: {Few}-{Viewpoint} {Surface} {Reconstruction} {Using} {Single}-{Photon} {Lidar}, 2024.
\newblock arXiv:2408.12191.

\bibitem[Malik et~al.(2023)Malik, Mirdehghan, Nousias, Kutulakos, and Lindell]{malik_transient_2023}
Anagh Malik, Parsa Mirdehghan, Sotiris Nousias, Kyros Kutulakos, and David Lindell.
\newblock Transient {Neural} {Radiance} {Fields} for {Lidar} {View} {Synthesis} and {3D} {Reconstruction}.
\newblock \emph{Advances in Neural Information Processing Systems}, 36:\penalty0 71569--71581, 2023.

\bibitem[Malik et~al.(2024)Malik, Juravsky, Po, Wetzstein, Kutulakos, and Lindell]{malik2024flying}
Anagh Malik, Noah Juravsky, Ryan Po, Gordon Wetzstein, Kiriakos~N Kutulakos, and David~B Lindell.
\newblock Flying with photons: Rendering novel views of propagating light.
\newblock In \emph{European Conference on Computer Vision}, pages 333--351. Springer, 2024.

\bibitem[Microelectronics()]{VL6180X_}
ST Microelectronics.
\newblock \emph{VL6180X Proximity and Ambient Light Sensing Module Datasheet}.
\newblock ST Microelectronics.

\bibitem[Morimoto et~al.(2020)Morimoto, Ardelean, Wu, Ulku, Antolovic, Bruschini, and Charbon]{morimoto_megapixel_2020}
Kazuhiro Morimoto, Andrei Ardelean, Ming-Lo Wu, Arin~Can Ulku, Ivan~Michel Antolovic, Claudio Bruschini, and Edoardo Charbon.
\newblock Megapixel time-gated {SPAD} image sensor for {2D} and {3D} imaging applications.
\newblock \emph{Optica}, 7\penalty0 (4):\penalty0 346, 2020.

\bibitem[Mu et~al.(2024)Mu, Sifferman, Jungerman, Li, Han, Gleicher, Gupta, and Li]{mu_towards_2024}
Fangzhou Mu, Carter Sifferman, Sacha Jungerman, Yiquan Li, Mark Han, Michael Gleicher, Mohit Gupta, and Yin Li.
\newblock Towards {3D} {Vision} with {Low}-{Cost} {Single}-{Photon} {Cameras}.
\newblock In \emph{2024 {IEEE}/{CVF} {Conference} on {Computer} {Vision} and {Pattern} {Recognition} ({CVPR})}, pages 5302--5311, Seattle, WA, USA, 2024. IEEE.

\bibitem[Niclass et~al.(2005)Niclass, Rochas, Besse, and Charbon]{niclass_design_2005}
C. Niclass, A. Rochas, P.-A. Besse, and E. Charbon.
\newblock Design and characterization of a {CMOS} 3-{D} image sensor based on single photon avalanche diodes.
\newblock \emph{IEEE Journal of Solid-State Circuits}, 40\penalty0 (9):\penalty0 1847--1854, 2005.
\newblock Conference Name: IEEE Journal of Solid-State Circuits.

\bibitem[Park et~al.(2019)Park, Patten, and Vincze]{park_pix2pose_2019}
Kiru Park, Timothy Patten, and Markus Vincze.
\newblock {Pix2Pose}: {Pixel}-{Wise} {Coordinate} {Regression} of {Objects} for {6D} {Pose} {Estimation}.
\newblock In \emph{2019 {IEEE}/{CVF} {International} {Conference} on {Computer} {Vision} ({ICCV})}, pages 7667--7676, 2019.
\newblock arXiv:1908.07433 [cs].

\bibitem[Pavlakos et~al.(2024)Pavlakos, Shan, Radosavovic, Kanazawa, Fouhey, and Malik]{pavlakos_reconstructing_2024}
Georgios Pavlakos, Dandan Shan, Ilija Radosavovic, Angjoo Kanazawa, David Fouhey, and Jitendra Malik.
\newblock Reconstructing {Hands} in {3D} with {Transformers}.
\newblock In \emph{2024 {IEEE}/{CVF} {Conference} on {Computer} {Vision} and {Pattern} {Recognition} ({CVPR})}, pages 9826--9836, Seattle, WA, USA, 2024. IEEE.

\bibitem[Romero et~al.(2017)Romero, Tzionas, and Black]{MANO:SIGGRAPHASIA:2017}
Javier Romero, Dimitrios Tzionas, and Michael~J. Black.
\newblock Embodied hands: Modeling and capturing hands and bodies together.
\newblock \emph{ACM Transactions on Graphics, (Proc. SIGGRAPH Asia)}, 36\penalty0 (6), 2017.

\bibitem[Ruget et~al.(2022)Ruget, Tyler, Mora-Martín, Scholes, Zhu, Gyongy, Hearn, McLaughlin, Halimi, and Leach]{ruget_pixels2pose_2022}
Alice Ruget, Max Tyler, Germán Mora-Martín, Stirling Scholes, Feng Zhu, Istvan Gyongy, Brent Hearn, Steve McLaughlin, Abderrahim Halimi, and Jonathan Leach.
\newblock {Pixels2Pose}: {Super}-{Resolution} {Time}-of-{Flight} {Imaging} for {3D} {Pose} {Estimation}.
\newblock In \emph{Imaging and {Applied} {Optics} {Congress} 2022 ({3D}, {AOA}, {COSI}, {ISA}, {pcAOP})}, page ITh5D.5, Vancouver, British Columbia, 2022. Optica Publishing Group.

\bibitem[Schwartz et~al.(2008)Schwartz, Charbon, and Shepard]{schwartz2008single}
David~Eric Schwartz, Edoardo Charbon, and Kenneth~L Shepard.
\newblock A single-photon avalanche diode array for fluorescence lifetime imaging microscopy.
\newblock \emph{IEEE journal of solid-state circuits}, 43\penalty0 (11):\penalty0 2546--2557, 2008.

\bibitem[Sifferman et~al.(2023)Sifferman, Wang, Gupta, and Gleicher]{sifferman_unlocking_2023}
Carter Sifferman, Yeping Wang, Mohit Gupta, and Michael Gleicher.
\newblock Unlocking the {Performance} of {Proximity} {Sensors} by {Utilizing} {Transient} {Histograms}.
\newblock \emph{IEEE Robotics and Automation Letters}, 8\penalty0 (10):\penalty0 6843--6850, 2023.

\bibitem[Sifferman et~al.(2024)Sifferman, Sun, Gupta, and Gleicher]{sifferman_using_2024}
Carter Sifferman, William Sun, Mohit Gupta, and Michael Gleicher.
\newblock Using a {Distance} {Sensor} to {Detect} {Deviations} in a {Planar} {Surface}.
\newblock \emph{IEEE Robotics and Automation Letters}, 9\penalty0 (10):\penalty0 8515--8522, 2024.
\newblock Conference Name: IEEE Robotics and Automation Letters.

\bibitem[{TechInsights}(2023)]{techinsights2023iphone}
{TechInsights}.
\newblock {iPhone 15 Pro Max Rear LiDAR Camera Process Flow Analysis}, 2023.
\newblock Accessed: 2025-03-07.

\bibitem[Tobin et~al.(2017)Tobin, Fong, Ray, Schneider, Zaremba, and Abbeel]{tobin2017domain}
Josh Tobin, Rachel Fong, Alex Ray, Jonas Schneider, Wojciech Zaremba, and Pieter Abbeel.
\newblock Domain randomization for transferring deep neural networks from simulation to the real world.
\newblock In \emph{2017 IEEE/RSJ international conference on intelligent robots and systems (IROS)}, pages 23--30. IEEE, 2017.

\bibitem[Tremblay et~al.(2023)Tremblay, Wen, Blukis, Sundaralingam, Tyree, and Birchfield]{tremblay_diff-dope_2023}
Jonathan Tremblay, Bowen Wen, Valts Blukis, Balakumar Sundaralingam, Stephen Tyree, and Stan Birchfield.
\newblock Diff-{DOPE}: {Differentiable} {Deep} {Object} {Pose} {Estimation}, 2023.
\newblock arXiv:2310.00463 [cs].

\bibitem[{Universal Robots}(2016)]{UR5Datasheet2016}
{Universal Robots}.
\newblock Ur5 technical specifications.
\newblock Online, 2016.
\newblock Accessed: 2025-02-28.

\bibitem[Vaswani et~al.(2017)Vaswani, Shazeer, Parmar, Uszkoreit, Jones, Gomez, Kaiser, and Polosukhin]{vaswani2017attention}
Ashish Vaswani, Noam Shazeer, Niki Parmar, Jakob Uszkoreit, Llion Jones, Aidan~N Gomez, {\L}ukasz Kaiser, and Illia Polosukhin.
\newblock Attention is all you need.
\newblock \emph{Advances in neural information processing systems}, 30, 2017.

\bibitem[Wen et~al.(2024)Wen, Yang, Kautz, and Birchfield]{wen_foundationpose_2024}
Bowen Wen, Wei Yang, Jan Kautz, and Stan Birchfield.
\newblock {FoundationPose}: {Unified} {6D} {Pose} {Estimation} and {Tracking} of {Novel} {Objects}.
\newblock In \emph{2024 {IEEE}/{CVF} {Conference} on {Computer} {Vision} and {Pattern} {Recognition} ({CVPR})}, pages 17868--17879, Seattle, WA, USA, 2024. IEEE.

\bibitem[Xiang et~al.(2018)Xiang, Schmidt, Narayanan, and Fox]{xiang_posecnn_2018}
Yu Xiang, Tanner Schmidt, Venkatraman Narayanan, and Dieter Fox.
\newblock {PoseCNN}: {A} {Convolutional} {Neural} {Network} for {6D} {Object} {Pose} {Estimation} in {Cluttered} {Scenes}, 2018.
\newblock arXiv:1711.00199 [cs].

\bibitem[Xin et~al.(2019)Xin, Nousias, Kutulakos, Sankaranarayanan, Narasimhan, and Gkioulekas]{xin_theory_2019}
Shumian Xin, Sotiris Nousias, Kiriakos~N. Kutulakos, Aswin~C. Sankaranarayanan, Srinivasa~G. Narasimhan, and Ioannis Gkioulekas.
\newblock A {Theory} of {Fermat} {Paths} for {Non}-{Line}-{Of}-{Sight} {Shape} {Reconstruction}.
\newblock In \emph{Proceedings of the IEEE/CVF conference on Computer Vision and Pattern Recognition}, pages 6800--6809, 2019.

\bibitem[Xu et~al.(2023)Xu, Yu, Xue, Ye, Yao, and Lu]{xu_visual-tactile_2023}
Wenqiang Xu, Zhenjun Yu, Han Xue, Ruolin Ye, Siqiong Yao, and Cewu Lu.
\newblock Visual-{Tactile} {Sensing} for {In}-{Hand} {Object} {Reconstruction}.
\newblock In \emph{2023 {IEEE}/{CVF} {Conference} on {Computer} {Vision} and {Pattern} {Recognition} ({CVPR})}, pages 8803--8812, Vancouver, BC, Canada, 2023. IEEE.

\bibitem[Young et~al.(2024)Young, Batagoda, Zhang, Dave, Pediredla, Negrut, and Raskar]{young_enhancing_2024}
Aaron Young, Nevindu~M. Batagoda, Harry Zhang, Akshat Dave, Adithya Pediredla, Dan Negrut, and Ramesh Raskar.
\newblock Enhancing {Autonomous} {Navigation} by {Imaging} {Hidden} {Objects} using {Single}-{Photon} {LiDAR}, 2024.
\newblock arXiv:2410.03555 [cs].

\bibitem[Zhou et~al.(2019)Zhou, Barnes, Lu, Yang, and Li]{zhou_continuity_2019}
Yi Zhou, Connelly Barnes, Jingwan Lu, Jimei Yang, and Hao Li.
\newblock On the {Continuity} of {Rotation} {Representations} in {Neural} {Networks}.
\newblock In \emph{2019 {IEEE}/{CVF} {Conference} on {Computer} {Vision} and {Pattern} {Recognition} ({CVPR})}, pages 5738--5746, 2019.
\newblock ISSN: 2575-7075.

\bibitem[Örnek et~al.(2025)Örnek, Labbé, Tekin, Ma, Keskin, Forster, and Hodan]{leonardis_foundpose_2025}
Evin~Pınar Örnek, Yann Labbé, Bugra Tekin, Lingni Ma, Cem Keskin, Christian Forster, and Tomas Hodan.
\newblock {FoundPose}: {Unseen} {Object} {Pose} {Estimation} with {Foundation} {Features}.
\newblock In \emph{Computer {Vision} – {ECCV} 2024}, pages 163--182. Springer Nature Switzerland, Cham, 2025.
\newblock Series Title: Lecture Notes in Computer Science.

\end{thebibliography}
}

\clearpage
\setcounter{figure}{0}
\setcounter{table}{0}
\setcounter{section}{0}
\setcounter{equation}{0}
\renewcommand{\thefigure}{\Alph{figure}}
\renewcommand{\thesection}{\Alph{section}}
\renewcommand{\thetable}{\Alph{table}}
\renewcommand{\theequation}{\Alph{equation}}
\setcounter{page}{1}
\maketitlesupplementary



In this supplementary material, we provide (1) a description of loss functions for training our feedforward model (\cref{sec:6d_pose_estimation_loss_fn}); (2) additional results on 6D pose estimation (\cref{sec:additional_6d_pose_experiments}); (3) an analysis of runtime and complexity (\cref{sec:runtime_complexity}); (4) experiments and discussion on sensor interference (\cref{sec:interference_test}); and (5) additional visualization of our results on 6D pose estimation (\cref{sec:additional_visualization}).

For sections, figures and equations, we use numbers (\eg, Sec.\ 1) to refer to the main paper and capital letters (\eg, Sec.\ A) to refer to this supplement.

\section{Training Loss of Feedforward Models} \label{sec:6d_pose_estimation_loss_fn}

\subsection{6D Pose Estimation}
As described in~\cref{subsec:implementation_details}, we utilize one of two losses to train the feedforward model depending on if the object is symmetrical. For non-symmetrical objects, we utilize a combination rotation, translation, and point matching loss. Given a ground truth object rotation $\mathbf{R}_\text{gt}$ (represented by the 6D representation proposed by \cite{zhou_continuity_2019}) and translation $\mathbf{t}_\text{gt}$. Given a set of 3D points $\mathbf{x}_i$ on the object, the loss of the predicted rotation $\mathbf{R}$ and translation $\mathbf{t}$ is given by:
\begin{equation*}
    \mathcal{L} = \lambda_r \mathcal{L}_{rot} + \lambda_t \mathcal{L}_{trans} + \lambda_p \mathcal{L}_{pm}
\end{equation*}
where the loss terms are given by
\begin{equation*}
\begin{split}
     &\mathcal{L}_{rot} = \|\mathbf{R} - \mathbf{R}_{\text{gt}} \|_1,\\
     &\mathcal{L}_{trans} = \|\mathbf{t} - \mathbf{t}_{\text{gt}} \|_1,\\
     &\mathcal{L}_{pm} = \frac{1}{N} \sum_{i=1}^{N} \| (\mathbf{R} \mathbf{x}_i + \mathbf{t}) - (\mathbf{R}_{\text{gt}} \mathbf{x}_i + \mathbf{t}_{\text{gt}}) \|_2
\end{split}   
\end{equation*}
We set $\lambda_r = 1.0$, $\lambda_t = 0.5$, $\lambda_p = 0.1$ for our experiments.

For symmetric objects, we use ADD-S loss introduced in \cite{xiang_posecnn_2018}, where $\mathcal{X}$ represents the set of object points:
\begin{equation*}
    \mathcal{L}_{ADD-S} = \frac{1}{N} \sum_{i=1}^{N} \min_{\mathbf{x}_j \in \mathcal{X}} \| (\mathbf{R} \mathbf{x}_i + \mathbf{t}) - (\mathbf{R}_{\text{gt}} \mathbf{x}_j + \mathbf{t}_{\text{gt}}) \|_2
\end{equation*}

\subsection{Spherical Object Recovery}
For spherical object recovery (\cref{subsec:extension_spheres}), the scene is parameterized by the center point $\mathbf{c} \in \mathbb{R}^3$ and diameter $d$. Our loss function is a simple combination of error in the two components:
\begin{equation*}
    \mathcal{L} = \| \mathbf{c} - \mathbf{c}_{\text{gt}}\| + \lambda |d - d_{\text{gt}}|
\end{equation*}
We set $\lambda = 1$ for our experiments.

\subsection{Human Hand Pose Estimation}
For hand pose estimation (\cref{subsec:extension_hand}), we predict the MANO model~\cite{MANO:SIGGRAPHASIA:2017} shape parameters $\boldsymbol{\beta}$, pose parameters $\boldsymbol{\theta}$, global 3D rotation $\mathbf{R}$ (represented by the 6D representation proposed by \cite{zhou_continuity_2019}), and global 3D translation $\mathbf{t}$. The loss for a given prediction is given by:
\begin{equation*}
\begin{split}
    &\mathcal{L} = \lambda_s\mathcal{L}_{shape} + \lambda_p\mathcal{L}_{pose} + \lambda_r \mathcal{L}_{rot}\\
    &+\lambda_t \mathcal{L}_{trans} + \lambda_j\mathcal{L}_{joint} + \lambda_v\mathcal{L}_{vertex}\\
\end{split}
\end{equation*}
where the loss terms are given by
\begin{equation*}
\begin{split}
     &\mathcal{L}_{shape} = \|\boldsymbol{\beta} - \boldsymbol{\beta}_{\text{gt}} \|_1,\\
     &\mathcal{L}_{pose} = \|\boldsymbol{\theta} - \boldsymbol{\theta}_{\text{gt}} \|_1,\\
     &\mathcal{L}_{rot} = \|\mathbf{R} - \mathbf{R}_{\text{gt}} \|_1,\\
     &\mathcal{L}_{trans} = \|\mathbf{t} - \mathbf{t}_{\text{gt}} \|_1,\\
     &\mathcal{L}_{j} = \|(\mathbf{R}\mathcal{M}_{j}(\boldsymbol{\beta}, \boldsymbol{\theta}) + \mathbf{t}) -  (\mathbf{R}_{\text{gt}}\mathcal{M}_{j}(\boldsymbol{\beta}_{\text{gt}}, \boldsymbol{\theta}_{\text{gt}}) + \mathbf{t}_{\text{gt}})\|_2,\\
      &\mathcal{L}_{v} = \|(\mathbf{R}\mathcal{M}_{v}(\boldsymbol{\beta}, \boldsymbol{\theta}) + \mathbf{t}) -  (\mathbf{R}_{\text{gt}}\mathcal{M}_{v}(\boldsymbol{\beta}_{\text{gt}}, \boldsymbol{\theta}_{\text{gt}}) + \mathbf{t}_{\text{gt}})\|_2\\
 \end{split}
\end{equation*}
Where $\mathcal{M}_{j}$ is the MANO model that outputs joint keypoint positions, and $\mathcal{M}_{v}$ is the MANO model that outputs mesh vertex positions.
We set $\lambda_s = 0.1$, $\lambda_p = 0.1$, $\lambda_r = 1.0$, $\lambda_t = 1.0, \lambda_j = 0.1, \lambda_v = 0.1$ for our experiments.

\section{Additional 6D Pose Estimation Experiments}\label{sec:additional_6d_pose_experiments}
\subsection{Data Visualization}
\begin{figure}[th]
    \centering
    \includegraphics[width=0.95\linewidth]{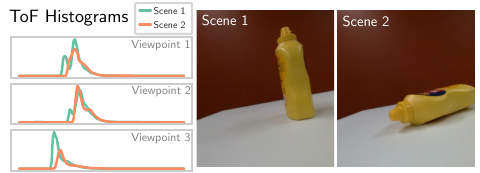} \vspace{-0.5em}
    \caption{Transient histograms from multiple viewpoints alongside corresponding 3D scenes.}
    \label{fig:histogram_example}
    \vspace{-0.5em}
\end{figure}

We visualize the transient histograms captured by multiple, distributed ToF sensors across two different 3D scenes in \cref{fig:histogram_example}.
The measurement has a complex relationship with scene geometry. We aim to solve the inverse problem (multi-view transient histogram $\rightarrow$ geometry) for simple parametric scenes.

\subsection{Fine-Tuning on Real Data}\label{subsec:fine_tuning}
We investigate the effects of fine-tuning our feedforward model on real data. To do so, we capture 80 additional measurements of the matte white ``2'' object used in prior experiments, and fine-tune the model trained on simulated data on these measurements. We leave the refiner unmodified.

The results of the fine-tuning experiment are presented in \cref{tab:6d_pose_finetune}. We see a significant improvement in the performance of the feedforward network. We also see a significant improvement in the result after refinement due to the improved starting estimate from the feedforward network. These results are encouraging as they indicate that a minimal amount of real-world data could improve the performance of our method.

\begin{table}[t]
\centering
\footnotesize
\begin{tabular}{lcccc}
\toprule
& \multicolumn{2}{c}{\textbf{AUC-ADD} ($\uparrow$)} \\
\textbf{Training Data} & \textbf{Feedforward} & \textbf{FF + Refiner} \\
\midrule
Fully Sim.        & 74.67   & 83.47  \\
Finetune on Real  & \textbf{86.16}   & \textbf{90.36}  \\
\bottomrule
\end{tabular}
\caption{Results of fine-tuning the 6D pose estimation method on real data, over 25 measurements of the ``2'' object.}\label{tab:6d_pose_finetune}
\vspace{-1em}
\end{table}

\subsection{Varying Scene Reflectance}
\label{subsec:vary_object_materials}
\begin{figure}
    \centering
    \includegraphics[width=0.90\linewidth]{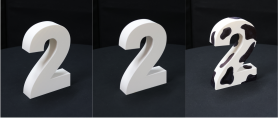}
    \caption{``2'' objects with different reflectance properties used in the varying scene reflectance experiment (\cref{subsec:vary_object_materials}. From left to right: matte white, glossy white, and spotted black and white.}
    \label{fig:two_materials}
    \vspace{-0.5em}
\end{figure}

The transient is a product of scene geometry \textit{and} reflectance, so scenes of varying reflectance could affect the performance of our method. We conduct a systematic test in which we modify the reflectance properties of the 3D printed digit ``2'' and the tabletop surface. We test ``2'' objects with three surface finishes, as shown in~\cref{fig:two_materials}. We test two table materials: matte white and matte black.

The results of varying surface properties are presented in \cref{tab:6d_pose_materials}. A modest decline in performance is observed with the glossy white object and the matte black tabletop, while a significant drop in performance occurs with the spotted black-and-white object. We attribute this drop to the fact that the spotted object has strong low-frequency variations in albedo across the surface. This sort of albedo variation is not included in our domain randomization when generating simulated data, nor is it able to be modeled by our refiner.

\begin{table}[t]
\centering
\footnotesize
\begin{tabular}{llcc}
\toprule
& & \multicolumn{2}{c}{\textbf{AUC-ADD} (\textuparrow)} \\
\textbf{Obj. Material} & \textbf{Table Material} & \textbf{FF} & \textbf{FF + Refiner}  \\
\midrule
Matte White & Matte White        & 74.67      & 83.47 \\
Matte White & Matte Black        & 66.59      & 79.55 \\
Glossy White & Matte White       & 69.86      & 77.74 \\
Spotted B/W & Matte White        & 50.46      & 61.49 \\
\bottomrule
\end{tabular}
\caption{6D Pose Estimation of the ``2'' object with varying object and tabletop surface reflectance.}\label{tab:6d_pose_materials}
\vspace{-0.5em}
\end{table}

\subsection{Varying Ambient Light}
\label{subsec:vary_lighting_conditions}
We evaluate the performance of our method under varying levels of ambient lighting in~\cref{tab:6d_pose_light_levels}, on a new set of 10 captures of the ``2'' object at each light level. We see consistent performance in darkness ($<$0.1 lux) and the same indoor lights as used in other captures (300 lux), but a heavy falloff in performance under a very bright halogen spotlight (3000 lux), which emits high amounts of infrared light, leading to a high DC offset in the transient histogram. This performance drop is expected as we assume negligible ambient light in both synthetic data generation and the refiner. Future work could aim to alleviate this problem by including ambient light level in domain randomization to make the feedforward network more robust, pre-processing histograms to mitigate the effect of ambient light, and/or optimizing for ambient light level in the refiner.

\begin{table}[t]
\centering
\footnotesize
\begin{tabular}{lcc}
\toprule
\textbf{Ambient Light Level} & \textbf{AUC-ADD}(\textuparrow) & \textbf{AUC-ADD-S}(\textuparrow)\\
\midrule
$<0.1$ lux & 65.69 & 90.47 \\
300 lux  & 72.45 & 93.10 \\
3000 lux (heavy IR) & 25.45 & 26.53 \\
\bottomrule
\end{tabular}
\caption{6D Pose Estimation of the ``2'' object under varying levels of ambient illumination.}\label{tab:6d_pose_light_levels}\vspace{-1em}
\end{table}

\subsection{Sensitivity to Inaccuracy in Sensor Pose}
\label{subsec:sensitivity_to_inaccuracy_in_sensor_pose}
\begin{figure}
    \centering
    \includegraphics[width=0.95\linewidth]{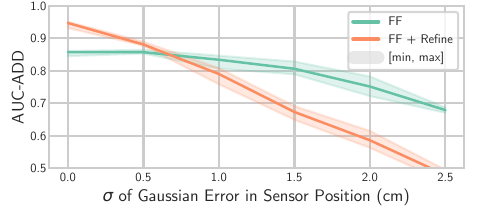}\vspace{-0.5em}
    \caption{Effect of adding Gaussian error to sensor poses on the ``2'' pose estimation task before feeding into our method.}
    \label{fig:sensor_pose_ablation}
    \vspace{-0.5em}
\end{figure}

When generating synthetic data to train our method, we add random Gaussian noise to the simulated sensor position to increase robustness to real-world inaccuracies in sensor position. We perform a simulated experiment to test this robustness, the results of which are shown in \cref{fig:sensor_pose_ablation}. Both the feedforward model and refiner are robust to modest variations in sensor pose ($<1$cm), which are likely achievable in realistic settings. We find that the feedforward method is more robust to variations than the refiner, and when there is high variation in sensor pose, foregoing the refiner leads to higher accuracy in the recovered object pose.

\subsection{Sensor Model Ablation Study} \label{subsec:sensor_model_ablation}
We perform an ablation study over key components of our sensor model as described in~\cref{subsec:sensor_model}. We consider the following variants: 

\begin{enumerate}
    \item \textbf{Full:} The full sensor model as described in~\cref{subsec:sensor_model} and used for all previous experiments.
    \item \textbf{Idealized Jitter Kernel:} The jitter kernel $\bs$ is replaced by a Dirac delta function at the location of the peak of $\bs$.
    \item \textbf{Inaccurate Bin Size:} The temporal bin size $\Delta t$ of the transient histogram is $\sim$10\% smaller than as calibrated (from 1.38cm to 1.2cm).
    \item \textbf{Inaccurate FoV Size:} The angular size of the FoV is incorrect by $\sim$ 20\%, increasing from $32^\circ$ to $38^\circ$. Additionally, the intensity map $I(\bw)$ is replaced with a constant function.
\end{enumerate}

For each variant, we train a feedforward model on synthetic data generated with the ablated sensor model, and use the same ablated sensor model in our refiner. Results over the 25-pose ``2'' digit dataset are shown in~\cref{tab:sensor_model_ablation}. The results demonstrate that each of these aspects of sensor modeling are important to achieve good performance.

\begin{table}[t]
\centering
\footnotesize
\begin{tabular}{lcc}
\toprule
& \multicolumn{2}{c}{\textbf{AUC-ADD (AUC-ADD-S)} ($\uparrow$)} \\
\textbf{Number of Views} & \textbf{Feedforward} & \textbf{FF+Refiner} \\
\midrule
5 Pixels (Views)                  & 74.79 (90.34)            & 74.80 (90.34)                                                 \\
10 Pixels (Views)                   & 78.29 (90.57)            & 78.07 (90.63)                                                \\
15 Pixels (Views)                   & 84.65 (91.27)        & 84.79 (91.66)                                            \\
25 Pixels  (Views)                  & 87.48 (91.51)        & 87.40 (91.42)                                                  \\
50 Pixels  (Views)                  & 90.54 (94.33)            & 90.87 (94.59)                                                   \\
100 Pixels  (Views)                  & 91.47 (94.58)           & 91.49 (94.37)                                                  \\
\bottomrule
\end{tabular}
\caption{6D Pose Estimation with Different Numbers of Views.}\label{tab:6d_pose_views}
\vspace{-0.5em}
\end{table}


\begin{table}[t]
\centering
\footnotesize
\begin{tabular}{lccc}
\toprule
& \multicolumn{2}{c}{\textbf{AUC-ADD} ($\uparrow$)} \\
\textbf{Ablation} & \textbf{Feedforward} & \textbf{FF + Refiner}\\
\midrule
Full Model                     & 73.67      & 83.47  \\
Idealized Jitter Kernel   & 22.64      & 24.50  \\
Incorrect Bin Size        & 49.98      & 33.23  \\
Incorrect FoV             & 29.70      & 44.22  \\
\bottomrule
\end{tabular}
\caption{Results of 6D Pose Estimation under varying sensor model ablations, over a dataset of 25 captures of the ``2'' object.}
\label{tab:sensor_model_ablation}
\vspace{-1em}
\end{table}

\begin{figure}
    \centering
    \includegraphics[]{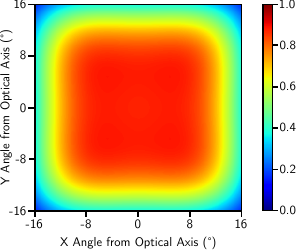}
    \caption{Visualization of the laser intensity function $I(\bw)$ that we use for the TMF8820 sensor, as given by~\cref{eqn:laser_intensity}. We set $K_1 = 0.88$, $K_2 = -3.16$, $K_3 = 250.51$.}
    \label{fig:laser_intensity}
    \vspace{-1em}
\end{figure}

\section{Runtime and Complexity Analysis}\label{sec:runtime_complexity}
While our method foregoes some computation performed by traditional methods (\eg peak finding and ICP), it is replaced by relatively costly neural network inference and iterative pose refinement. Therefore we do not foresee efficiency improvements compared to point cloud-based methods. One feed-forward pass of our network takes $\sim$4.8 ms. The (unoptimized) refiner takes $\sim$2 seconds. With attention paid to efficiency, refiner speed could likely be increased. The costs of both the forward pass and optimization scale linearly with the number of viewpoints.

\section{Test of Between-Sensor Interference}
\label{sec:interference_test}
In our prototype system, a single sensor is moved to multiple positions while the scene remains static. However many practical applications for our method may involve multiple sensors imaging the scene at the same time, which could lead to interference between sensors. We perform controlled experiments to investigate the effect of interference.

\subsection{Two Sensors Facing Each Other}
\begin{figure}
    \centering
    \begin{subfigure}{\columnwidth}
        \includegraphics[]{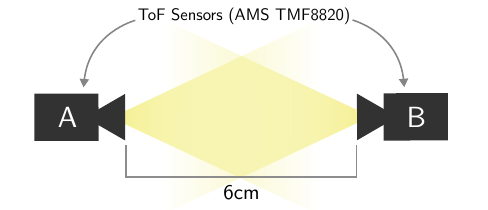}
        \caption{Sensor configuration for interference experiment 1.}
        \label{fig:interference_experiment_1_setup}
    \end{subfigure}
    \begin{subfigure}{\columnwidth}
        \includegraphics[]{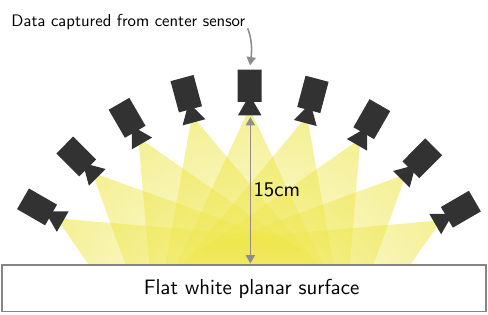}
        \caption{Sensor configuration for interference experiment 2.}
        \label{fig:interference_experiment_2_setup}
    \end{subfigure}
    \caption{Sensor configurations used for interference experiments.}
    \vspace{-1em}
\end{figure}

\begin{figure}
    \centering
    \begin{subfigure}{\columnwidth}
        \includegraphics[]{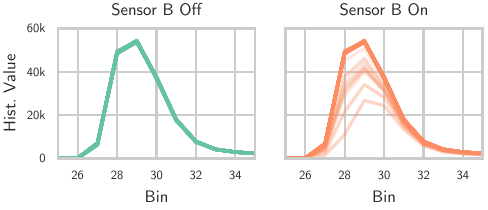}
        \caption{Raw histograms}
        \vspace{1em}
    \end{subfigure}
    \begin{subfigure}{\columnwidth}
        \includegraphics[]{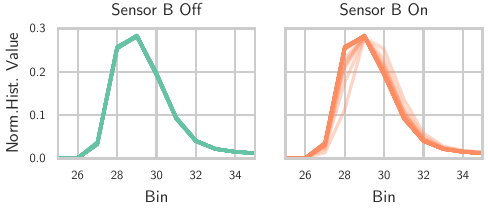}
        \caption{Histograms normalized to have a sum of 1.}
    \end{subfigure}
    \caption{Comparison of the histograms captured in interference experiment 1. Each plot shows 128 sensor measurements overlaid. About 90\% of samples in the right column exhibit no interference artifacts, comprising the dark orange lines.}
    \label{fig:interference_experiment_1a_results}
\end{figure}

We position two AMS TMF8820 sensors facing directly at each other at a distance of 6cm, as illustrated in~\cref{fig:interference_experiment_1_setup}. We compare measurements captured by sensor A between two conditions: sensor B on and sensor B off. The raw and normalized histograms for both conditions are shown in~\cref{fig:interference_experiment_1a_results}. We find that the operation of sensor B causes an effect in the histogram captured by sensor A $\sim10\%$ of the time. Even after normalization, the effect is still present. This effect appears similar to the effect caused by ambient light~\cite{gupta_photon-flooded_2019}, and is consistent with what we would expect to see if sensor B's light source is not correlated with the light source of sensor A; \ie, because the laser pulse trains of the two sensors are not synchronized, sensor B's operation leads to photons arriving uniformly at any time relative to sensor A's pulse train, just as ambient light arrives uniformly.

\begin{figure}
    \centering
    \includegraphics[]{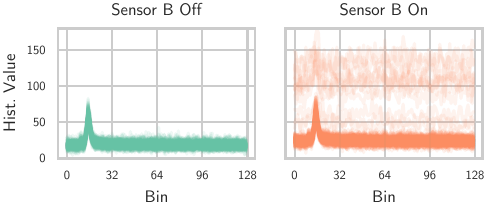}\vspace{-0.5em}
    \caption{Comparison on the histograms captured in interference 1, with the light source of sensor A covered. Each plot shows 128 sensor measurements overlaid. About 90\% of the samples in the right column exhibit no interference artifacts, comprising the dark orange line.}\label{fig:interference_experiment_1b_results}
\end{figure}

\begin{figure}[t]
    \centering
    \begin{subfigure}{\columnwidth}
        \includegraphics[]{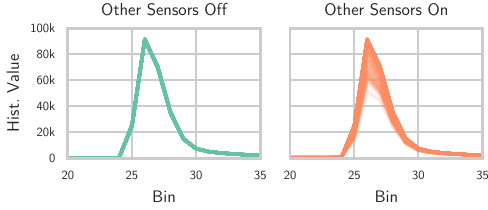}
        \caption{Raw histograms}
        \vspace{1em}
    \end{subfigure}
    \begin{subfigure}{\columnwidth}
        \includegraphics[]{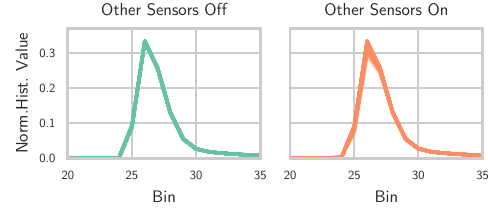}
        \caption{Histograms normalized to have a sum of 1}
        \vspace{1em}
    \end{subfigure}
    \begin{subfigure}{\columnwidth}
        \includegraphics[]{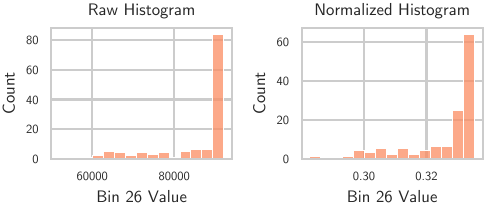}
        \caption{Histogram of values of bin 26 (the peak) for the "Other Sensors On" condition. The bin values are grouped together in about 75\% of measurements.}
    \end{subfigure}
    \caption{Comparison of the histograms captured in interference experiment 2.} \label{fig:interference_experiment_2_results}
\end{figure}

To further validate this hypothesis, we perform another test using the same sensor configuration in which the laser light source of sensor A is covered, so that only ambient light and the effect of sensor B are captured by sensor A. The results of this experiment are shown in~\cref{fig:interference_experiment_1b_results}. In this case we can clearly see interference manifest as a DC offset in the captured histogram, again matching the signature of ambient light.

\subsection{Nine Sensors Imaging a Plane}
We perform a second experiment in which nine sensors are all operating simultaneously and imaging the same portion of a planar surface. The experimental setup is illustrated in~\cref{fig:interference_experiment_2_setup}. We position the sensors such that the centers of their optical axes each intersect with a planar surface at the same point, and record data only from the center sensor. Again, we compare between two conditions: the other 8 sensors on, and the other 8 sensors off. The results of this experiment are shown in~\cref{fig:interference_experiment_2_results}. We see the same effect as in the previous experiment, but with a slightly higher occurrence rate of $\sim25\%$.

\subsection{Discussion: Between-Sensor Interference}
We have demonstrated that, at least for the AMS TMF8820 sensor, the effect of interference between sensors happens only occasionally even in the worst case. In practical scenarios, the rate of interference is likely to be quite low (\ie $<10\%$). Further, the effect of interference on the histogram appears to be similar to the effect of ambient light. Adjusting captured histograms to account for ambient light is a well-studied problem~\cite{gupta_photon-flooded_2019}, and it is likely that methods which are robust to changes in ambient light will be robust to between-sensor interference. While future applications should take interference into account, we believe it is unlikely to be a major obstacle for future deployments of distributed miniature ToF sensors.

\section{Visualization of 6D Pose Results}\label{sec:additional_visualization}

\begin{figure*}
    \centering
    \includegraphics{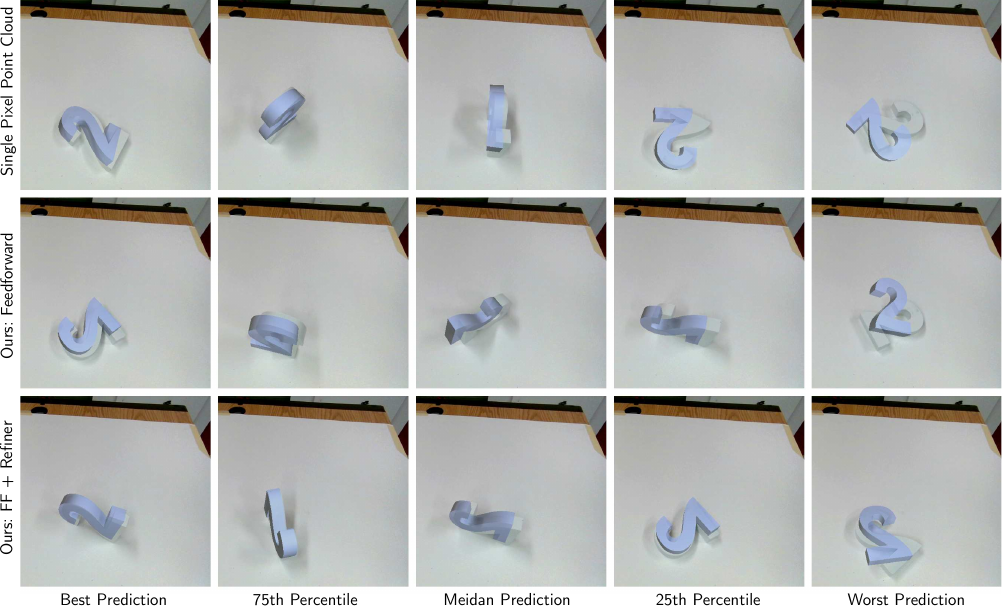}
    \caption{Visualization of results on the 3D printed ``two'' object.}
    \label{fig:two_grid}
\end{figure*}

\begin{figure*}
    \centering
    \includegraphics{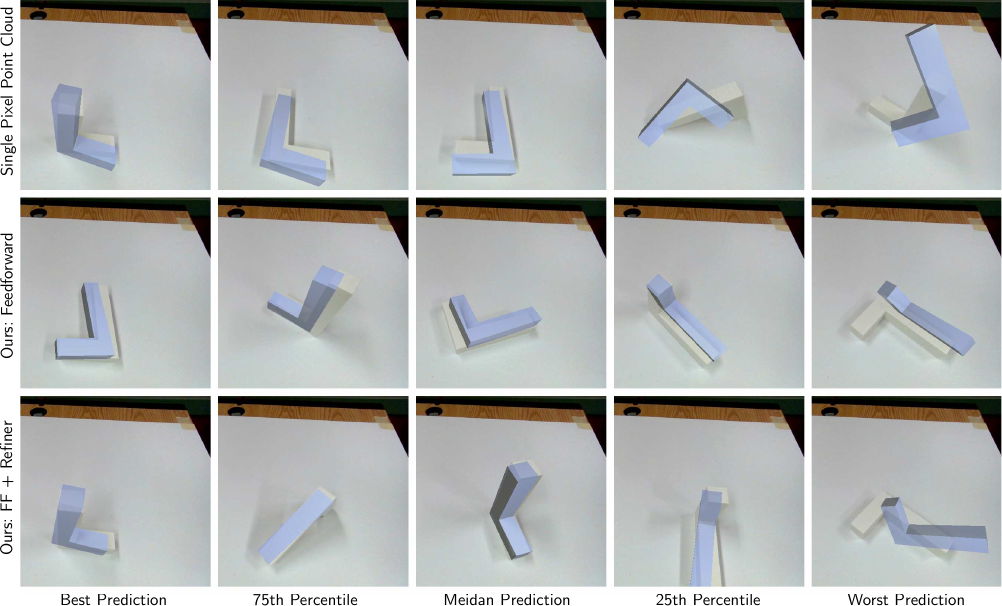}
    \caption{Visualization of results on the 3D printed ``L'' object.}
    \label{fig:L_grid}
\end{figure*}

\begin{figure*}
    \centering
    \includegraphics{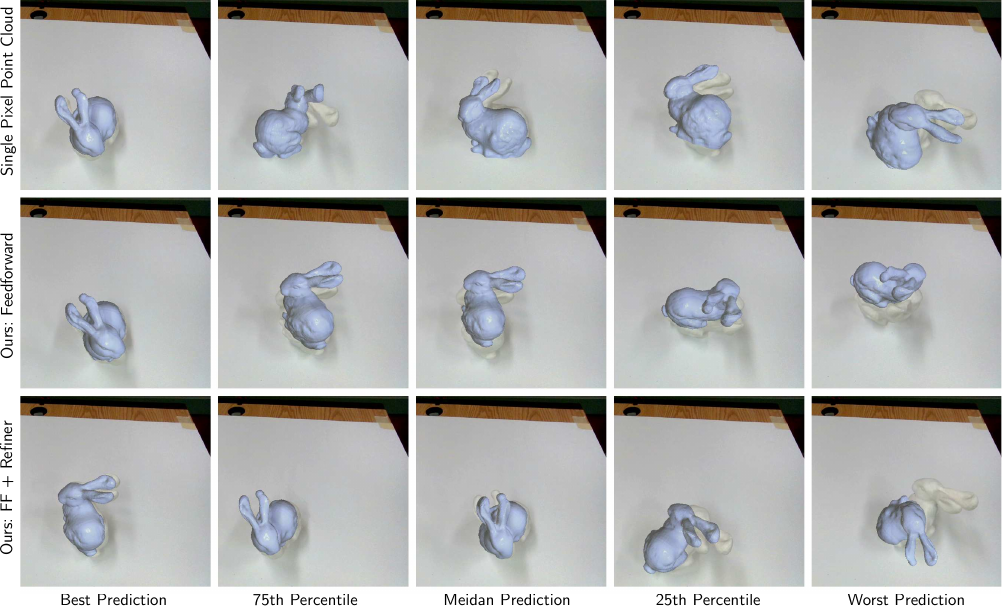}
    \caption{Visualization of results on the 3D printed ``bunny'' object.}
    \label{fig:bunny_grid}
\end{figure*}

\begin{figure*}
    \centering
    \includegraphics{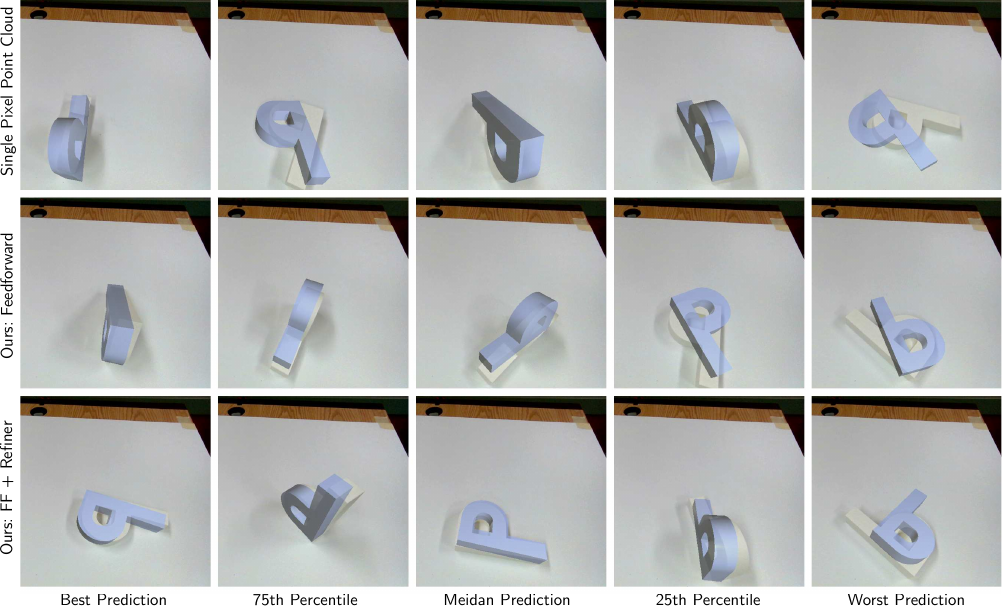}
    \caption{Visualization of results on the 3D printed ``P'' object.}
    \label{fig:P_grid}
\end{figure*}

\begin{figure*}
    \centering
    \includegraphics{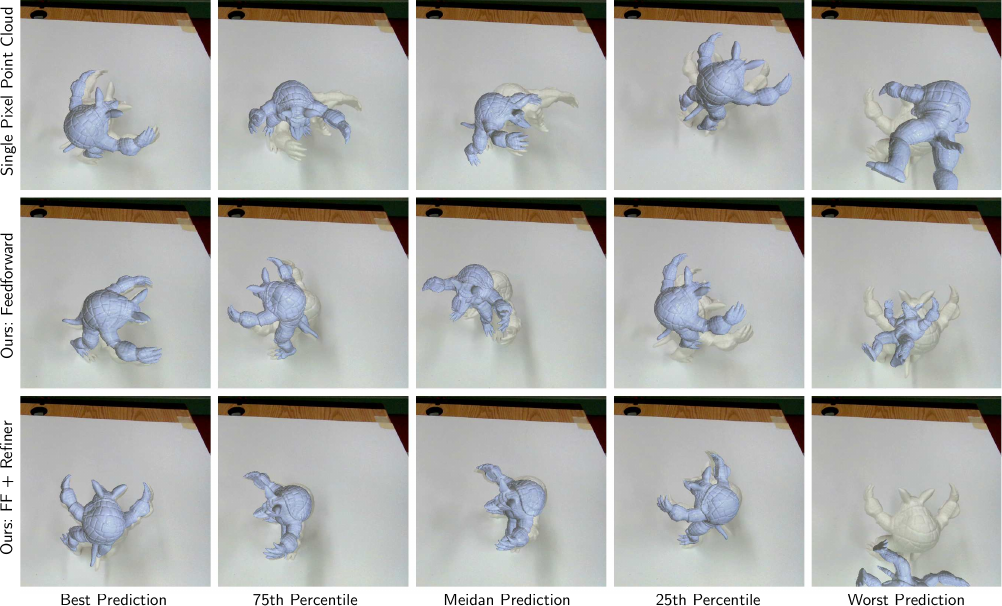}
    \caption{Visualization of results on the 3D printed ``armadillo'' object.}
    \label{fig:armadillo_grid}
\end{figure*}

\begin{figure*}
    \centering
    \includegraphics{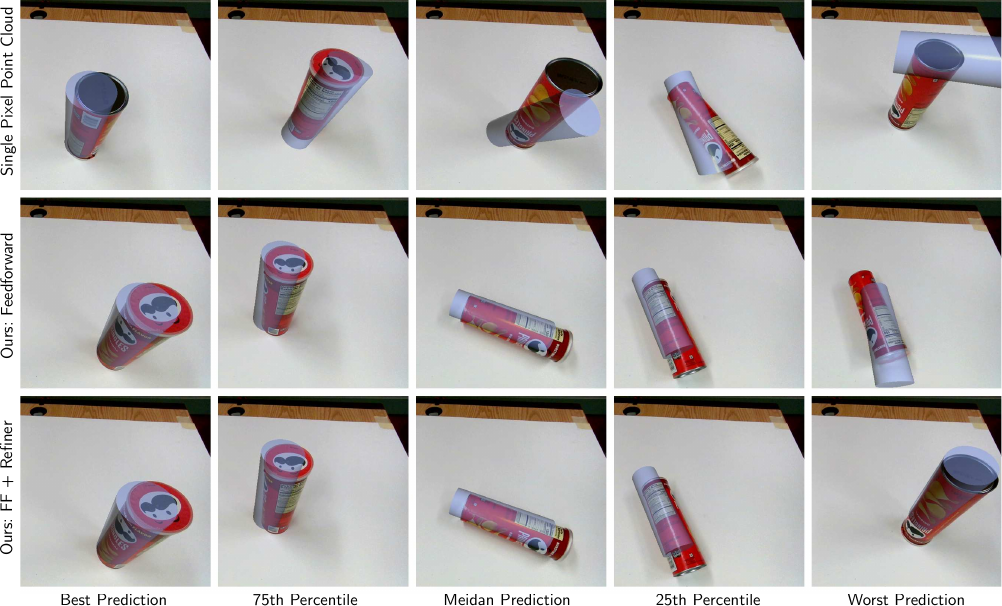}
    \caption{Visualization of results on the ``chips'' object from the YCB dataset.}
    \label{fig:chips_grid}
\end{figure*}

\begin{figure*}
    \centering
    \includegraphics{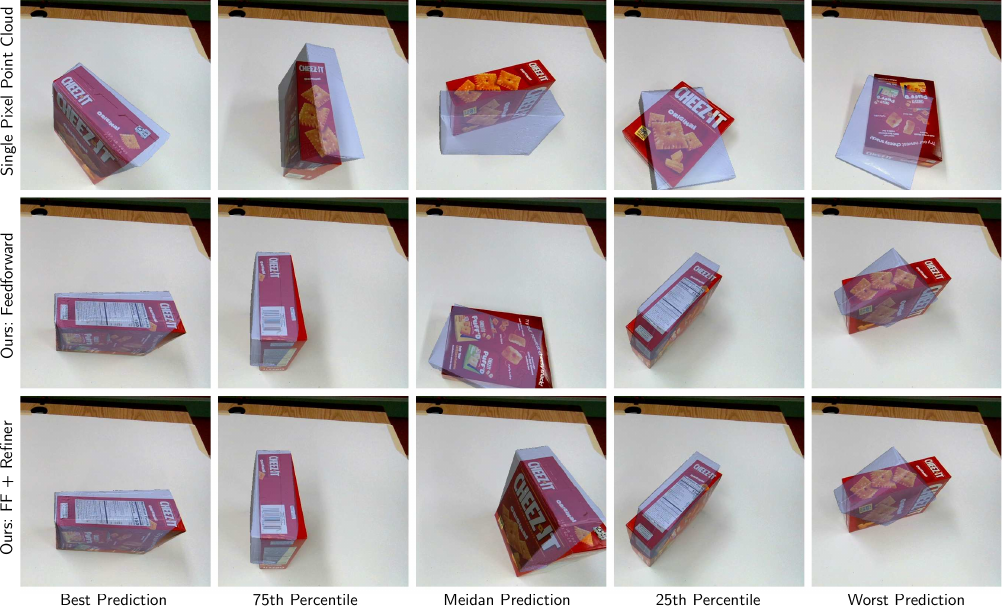}
    \caption{Visualization of results on the ``crackers'' object from the YCB dataset.}
    \label{fig:crackers_grid}
\end{figure*}

\begin{figure*}
    \centering
    \includegraphics{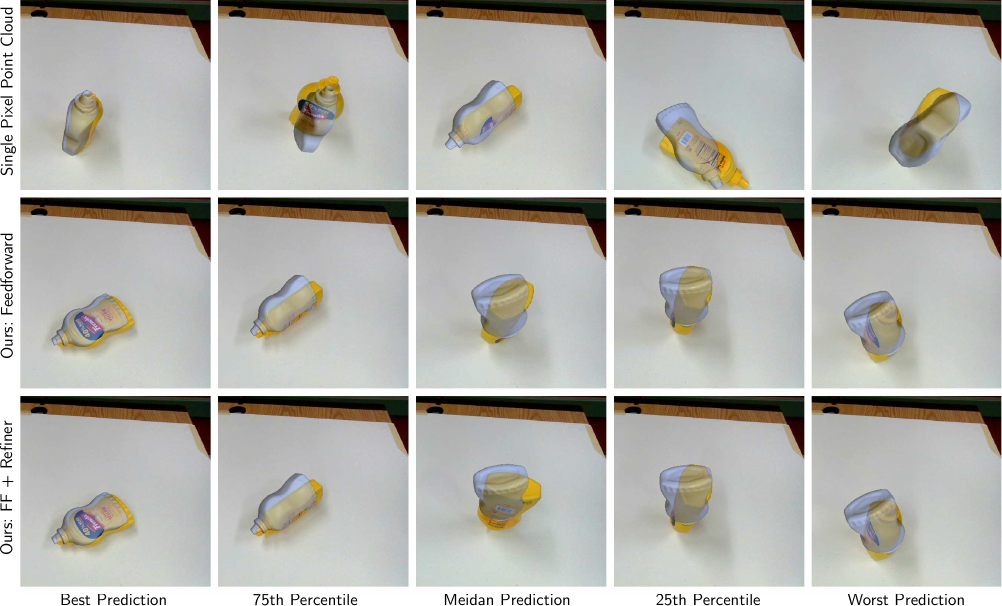}\vspace{-0.3em}
    \caption{Visualization of results on the ``mustard'' object from the YCB dataset.}
    \label{fig:mustard_grid}\vspace{-0.5em}
\end{figure*}

\begin{figure*}
    \centering
    \includegraphics{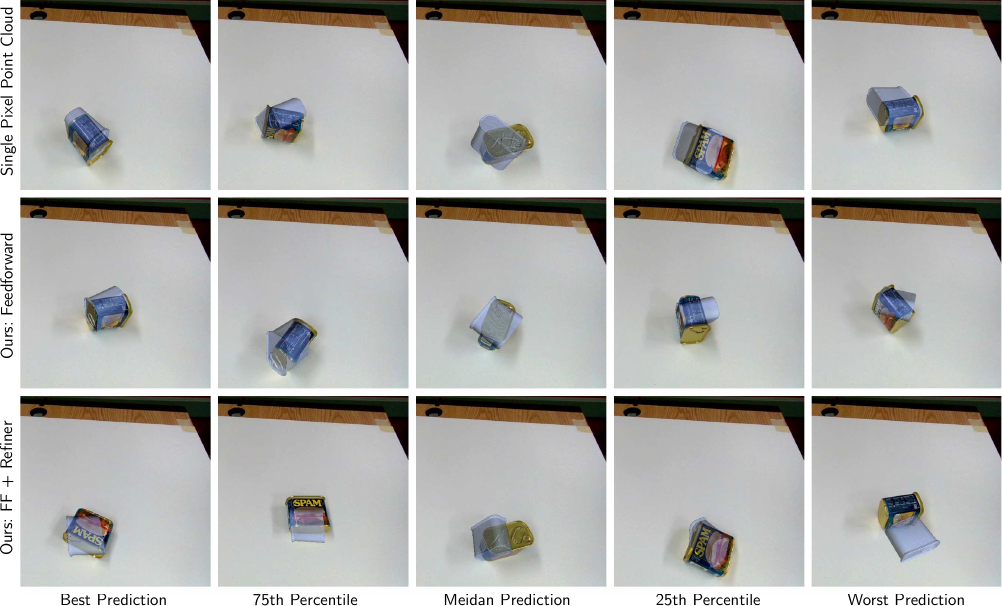}\vspace{-0.3em}
    \caption{Visualization of results on ``SPAM'' object from the YCB dataset. The SPAM is a failure case for our method due to its specular surface, small size, and many near-symmetries which make optimization difficult.}
    \label{fig:spam_grid}
\end{figure*}

We provide visualization of our results on 6D pose estimation in \Cref{fig:two_grid,fig:L_grid,fig:bunny_grid,fig:P_grid,fig:armadillo_grid,fig:chips_grid,fig:crackers_grid,fig:mustard_grid,fig:spam_grid}. 

\clearpage

\begin{figure*}
    \centering
    \includegraphics[]{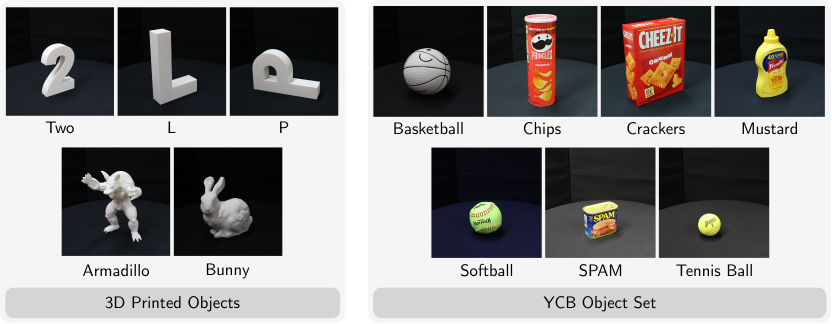}
    \caption{Objects used for 6D pose estimation experiments.}
    \label{fig:objects}
\end{figure*}

\end{document}